\documentclass[conference]{IEEEtran}
\IEEEoverridecommandlockouts
\usepackage{cite}
\usepackage{amsmath,amssymb,amsfonts}
\usepackage{algorithmic}
\usepackage{graphicx}
\usepackage{textcomp}
\usepackage{multirow}
\usepackage{cancel}
\usepackage{multicol}
\usepackage{comment}
\usepackage{caption}
\usepackage{subcaption}
\usepackage{pifont}
\usepackage[dvipsnames]{xcolor}
\usepackage{algorithm,algorithmic}
\def\BibTeX{{\rm B\kern-.05em{\sc i\kern-.025em b}\kern-.08em
    T\kern-.1667em\lower.7ex\hbox{E}\kern-.125emX}}
    
\makeatletter
\newcommand*\titleheader[1]{\gdef\@titleheader{#1}}
\AtBeginDocument{%
  \let\st@red@title\@title
  \def\@title{%
    \bgroup\normalfont\large\centering\@titleheader\par\egroup
    \vskip1.5em\st@red@title}
}
\makeatother    
    
\begin{document}
\title{
{Bidirectional Variational Autoencoders} \\
}

\author{\IEEEauthorblockN{Bart Kosko}
\IEEEauthorblockA{\textit{Department of Electrical and Computer Engineering} \\
\textit{University of Southern California}\\
Los Angeles, CA, USA \\
kosko@usc.edu}
\and
\IEEEauthorblockN{Olaoluwa Adigun}
\IEEEauthorblockA{\textit{Department of Electrical and Computer Engineering} \\
\textit{University of Southern California}\\
Los Angeles, CA, USA \\
adigun@usc.edu}
}

\maketitle
\begin{abstract}
We present the new bidirectional variational autoencoder (BVAE) network architecture.
The BVAE uses a single neural network both to encode and decode instead of an encoder-decoder network pair.
The network encodes in the forward direction and decodes in the backward direction through the \emph{same} synaptic web.
The network maximizes the BELBO or the \emph{bidirectional} evidence lower bound.
Simulations compared BVAEs and ordinary VAEs on the four image tasks of image reconstruction, classification, interpolation, and generation.
The image datasets included MNIST handwritten digits, Fashion-MNIST, CIFAR-10, and CelebA-64 face images.
The bidirectional structure of BVAEs cut the parameter count by almost 50\% and still slightly outperformed the unidirectional VAEs.
\end{abstract}

\begin{IEEEkeywords}
variational autoencoder, bidirectional backpropagation, directional likelihoods, ELBO, evidence lower bound
\end{IEEEkeywords}

\section{Directional Variational Autoencoders}

This paper introduces the new \emph{bidirectional} variational autoencoder (BVAE) network.
This architecture uses a single parametrized network for encoding and decoding.
It trains with the new bidirectional backpropagation algorithm that jointly optimizes the network's bidirectional likelihood \cite{adigun2016bidirectional, AdigunK20_v1}. 
The algorithm uses the same synaptic weights both to predict the target $y$ given the input $x$ and to predict the converse $x$ given $y$.
Ordinary or unidirectional VAEs use separate networks to encode and decode. \\


Unidirectional variational autoencoders (VAEs) are unsupervised machine-learning models that learn data representations \cite{SinhaD21, DavidsonFCKT18}.
They both learn and infer with directed probability models that often use intractable probability density functions \cite{KingmaW13}.
A VAE seeks the best estimate of the data likelihood $p(x|\theta)$ from samples $\{x^{(n)}\}_{n=1}^{N}$ if $x$ depends on some observable feature $z$ and if $\theta$ represents the system parameters.
The intractability involves marginalizing out the random variable $z$ to give the likelihood $p(x|\theta)$:
\begin{align}
p(x|\theta) &=  \mathbb{E}_{z|\theta} \big{[} p(x|z,\theta) \big{]} = \int_{z} p(x|z,\theta) \ p(z|\theta) \ dz .
\end{align}

\begin{figure*}[t]
  \centering
   \begin{subfigure}[b]{0.49\linewidth}
         \centering
         \includegraphics[width=\textwidth, height=0.57\textwidth]{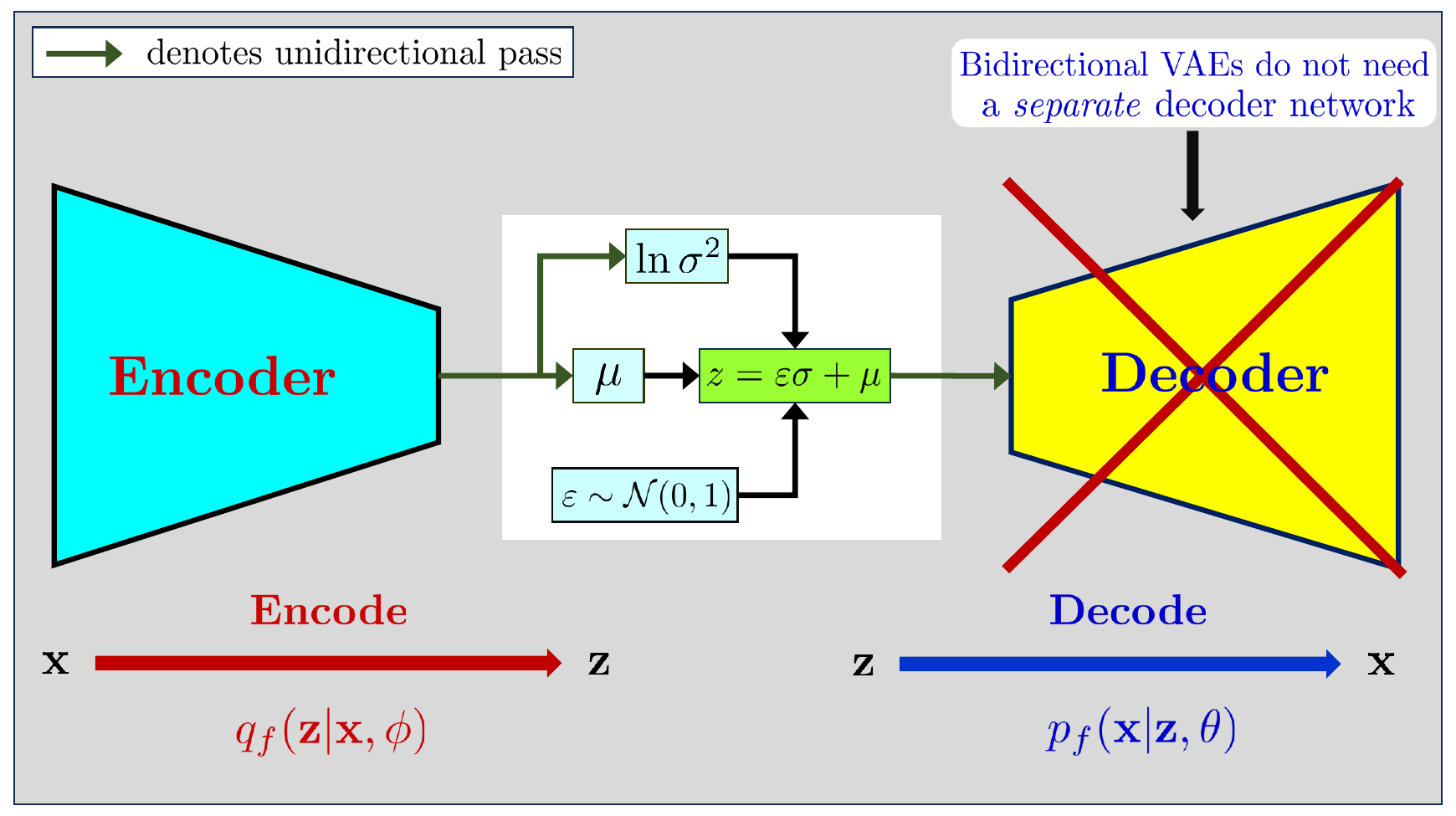}
         \caption{Unidirectional VAE architecture}
         \label{fig:uni_vae_arch}
   \end{subfigure}
   \hfill
   \begin{subfigure}[b]{0.49\linewidth}
         \centering
         \includegraphics[width=\textwidth]{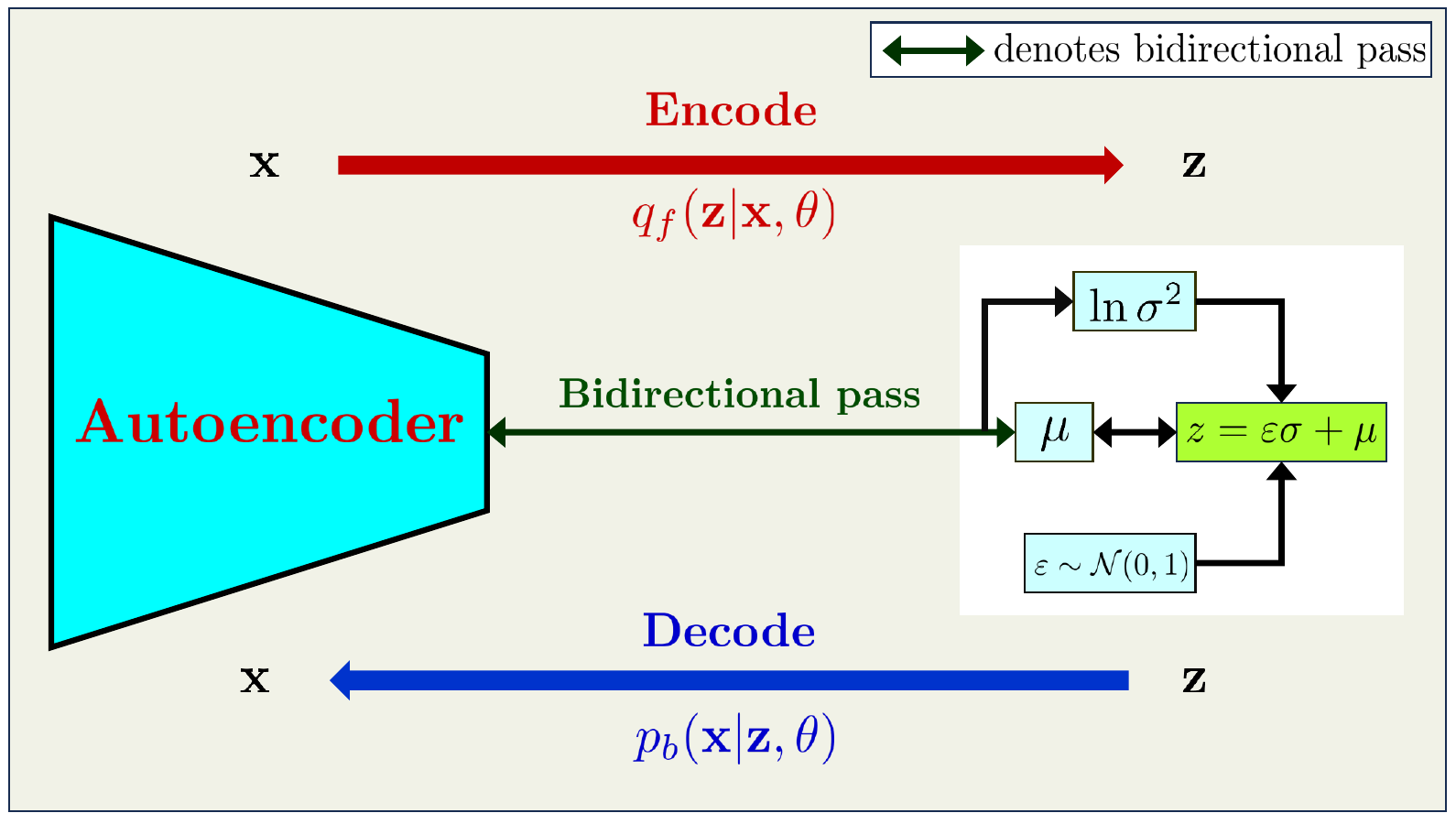}
         \caption{Bidirectional VAE architecture}
         \label{fig:bi_vae_arch}
   \end{subfigure}
   \caption{\small Bidirectional vs. unidirectional variational autoencoders:  Unidirectional VAEs use the forward passes of two separate networks for encoding and decoding.
   Bidirectional VAEs  encode on the forward pass and decode on the backward pass with the same synaptic weight matrices in both directions.
This cuts the number of tunable parameters roughly in half.
(a) The decoder network with parameter $\theta$ approximates $p(x|z, \theta)$ and the encoder network with parameter $\phi$ approximates $q(z|x, \theta)$.
(b) Bidirectional VAEs use the forward pass of a network with parameter $\theta$ to approximate $q(z|x,\theta)$ and the backward pass of the network to approximate $p(x|z,\theta)$.}
   \label{fig:main_image}
\end{figure*}

Kingma and Welling introduced VAEs to solve this computational problem \cite{KingmaW13}.
The VAE includes a new recognition (or encoding) model $q(z|x,\phi)$ that approximates the intractable likelihood $q(z|x,\theta)$.
The probability $q(z|x,\phi)$ represents a probabilistic \emph{encoder} while $p(x|z,\theta)$ represents a probabilistic \emph{decoder}.
These probabilistic models use two neural networks with different synaptic weights.
Figure \ref{fig:uni_vae_arch} shows the architecture of such a unidirectional VAE.
The recognition model doubles the number of parameters and the computational cost of this solution. 

The new bidirectional backpropagation (B-BP) algorithm trains and runs a neural network to run forwards and backwards by jointly maximizing the respective directional probabilities.  
This joint maximization allows the network to run backward from output code words to expected input patterns.
Running a unidirectionally trained network backwards just produces noise. 
B-BP jointly maximizes the forward likelihood $q_f(z|x,\theta)$ and backward likelihood $p_b(x|z,\theta)$ or the equivalent  sum of their respective log-likelihoods:
\begin{align}
\theta^{*} &= \underset{{\theta}}{\arg\max}\   q_f({z}|{x} , \theta) \ p_b({x}|{z} , \theta) \\
&=  \underset{{\theta}}{\arg\max}\  \underbrace{\ln  q_f({z}|{x} , \theta)}_{\substack{\text{\sf \scriptsize \textcolor{red}{Forward pass}}}} + \underbrace{\ln  p_b({x}|{z} , \theta)}_{\substack{\text{\sf \scriptsize \textcolor{blue}{Backward pass}}}} .
\end{align}

A BVAE approximates the intractable $q(z|x,\theta)$ with the forward likelihood $q_f(z|x,\theta)$.
Then the probabilistic \emph{encoder} is $q_f(z|x,\theta)$ and the probabilistic \emph{decoder} is $p_b(x|z,\theta)$.
So the two densities share the same parameter $\theta$ and there is no need for a separate network.
Figure \ref{fig:bi_vae_arch} shows the architecture of a BVAE. 

VAEs vary based on the choice of latent distribution, the method of training, and the use of joint modeling with other generative models, among other factors.
The $\beta$-VAE  introduced the adjustable hyperparameter $\beta$.
It balances the latent channel capacity of the encoder network and the reconstruction error of the decoder network \cite{higgins2016beta}.
It trains on a weighted sum of the reconstruction error and the Kullback-Leibler divergence $D_{\mathtt{KL}}\big{(} q(z|x,\phi) || p(z|\theta) \big{)}$.
The $\beta$-TCVAE (Total Correlation Variational Autoencoder) extends $\beta$-VAE to learning isolating sources of disentanglement \cite{chen2018isolating}.
A disentangled $\beta$-VAE modifies the $\beta$-VAE by progressively increasing the information capacity of the latent code while training \cite{Christopher2018}.  

Importance weighted autoencoders (IWAEs) use priority weights to derive a strictly tighter lower bound on the log-likelihood \cite{burda2015importance}. 
Variants of IWAE include  the partially importance weighted auto-encoder (PIWAE), the multiply importance weighted auto-encoder (MIWAE), and the combined importance weighted auto-encoder (CIWAE) \cite{rainforth2018tighter}.

Hyperspherical VAEs use a non-Gaussian latent probability density.
They use  a von Mises-Fisher (vMF) latent density that gives in turn a hyperspherical latent space \cite{DavidsonFCKT18}.
Other VAEs include the Consistency Regularization for Variational Auto-Encoder (CRVAE) \cite{SinhaD21}, the InfoVAE \cite{zhao2017infovae}, and the Hamiltonian VAE \cite{CateriniDS18} and so on.
All these VAEs use separate networks to encode and decode.  

{Vincent et.  al. \cite{JMLR_v11_vincent10a} suggests the use of tied weights in stacked autoencoder networks.
This is a form of constraint that parallels the working of restricted Boltzmann machines RBMs \cite{smolensky1986information} and thus a simple type of bidirectional associative memory or BAM \cite{kosko1988BAM}.
It forces the weights to be symmetric with $W^{-1}= W^T$
The building block here is a shallow network with no hidden layer \cite{hinton2006reducing,larochelle2008classification}.
They further suggest that combining this form of constraint with nonlinear activation would most likely lead to poor reconstruction error. 
}

Bidirectional autoencoders  BAEs \cite{adigun2023bidirectional} extend bidirectional neural representation to image compression and denoising.
BAEs differ from autoencoders with tied weight because it relaxes the constraint by extending the bidirectional assumption over the depth of a deep network.
BAEs differ from bidirectional VAEs because they do not require the joint optimization of the directional likelihoods.
This limits the generative capability of BAEs.

The next sections review ordinary VAEs and introduce the probabilistic BVAEs with the new B-BP algorithm.
 Section \ref{sec:exp_setup} compares them on the four standard image test datasets:  MNIST handwritten digits, Fashion-MNIST, CIFAR-10, and CelebA-64 datasets. 
We find that the BVAE models cut the tunable parameters roughly in half while still performing slightly better than the unidirectional VAEs.

\begin{figure*}[!t]
\centerline{\includegraphics[width=0.95\textwidth]{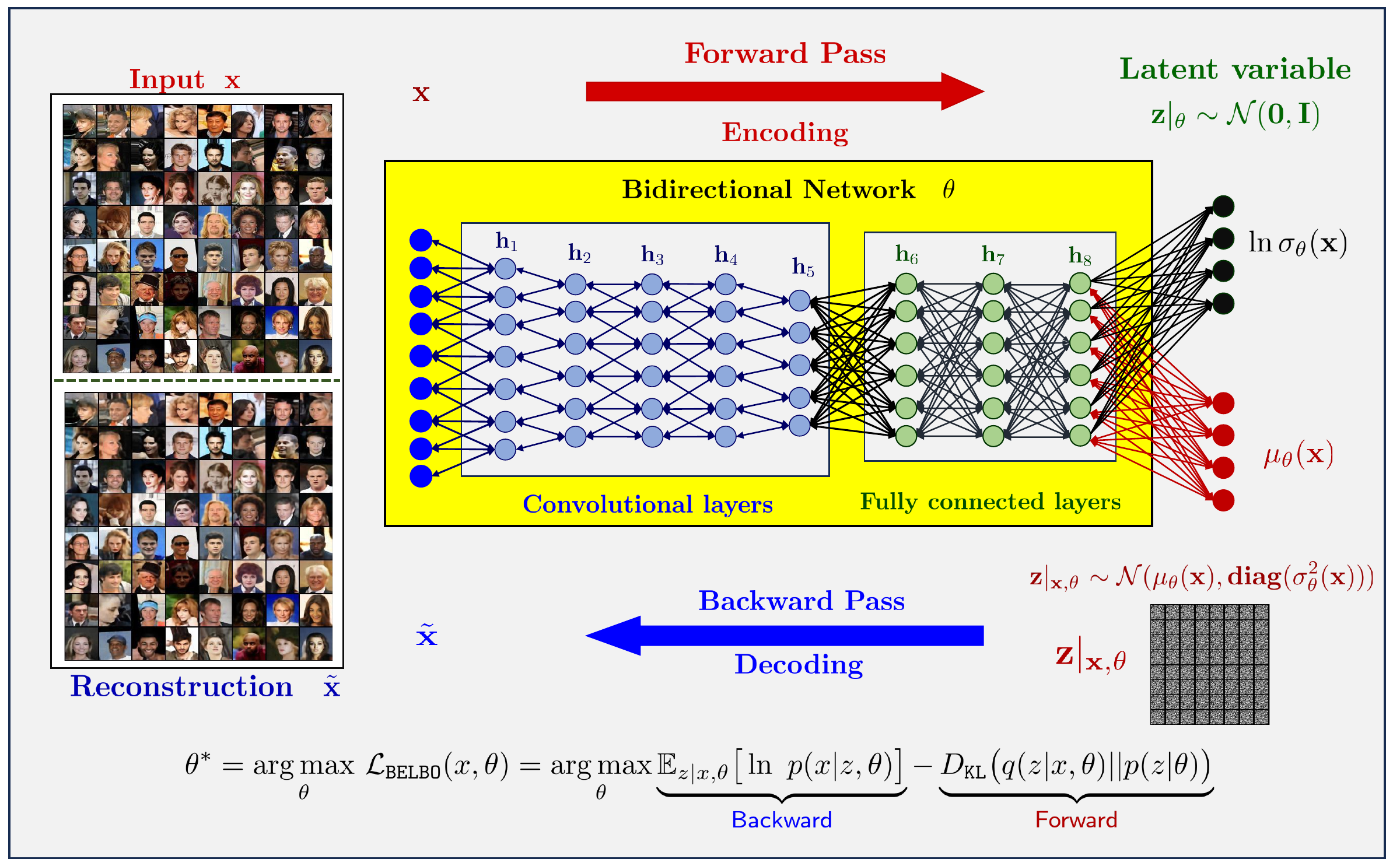}}
\caption{\small{
BELBO training of a bidirectional variational autoencoder with the bidirectional backpropagation algorithm.
BELBO maximization uses a single network for encoding and decoding.
The forward pass with likelihood $q_f(z|x, \theta)$ encodes the latent features.
The backward pass with likelihood $p_b(x|z, \theta)$ decodes the latent features over the same web of synapses.
}}
\label{fig:bidirectional_BP_training}
\end{figure*}

\section{Unidirectional Variational Autoencoders}
\label{sec:vae}

Let $p(x|\theta)$ be the data likelihood and $z$ denote the hidden variable.
The data likelihood simplifies as follows from the definition of conditional probability:
\begin{align}
p(x|\theta) &= \frac{p(x, z|\theta)}{q(z|x, \theta)} = \frac{p(x|z, \theta) p(z|\theta)}{q(z|x, \theta)}  \ .
\end{align}
The conditional likelihood $q(z|x,\theta)$ is intractable to solve so unidirectional VAEs introduce a new conditional likelihood  that represents the \emph{recognition} or \emph{encoding} model.
The term $q_f(z|x,\phi)$ represents the forward likelihood of the encoding network that approximates the intractable likelihood $q(z|x,\theta)$. 
We have 
\begin{align}
p(x|\theta) &= \frac{p(x|z, \theta) p(z|\theta)}{q(z|x, \theta)}  = \frac{p(x|z, \theta) p(z|\theta) q_f(z|x, \phi)}{q(z|x, \theta) q_f(z|x, \phi)} \ . 
\end{align}

The corresponding data log-likelihood $\ln p(x|\theta)$ is
\begin{align}
\ln p(x&|\theta) = \ln  \bigg{[} \frac{p(x|z, \theta) p(z|\theta) q_f(z|x, \phi)}{q(z|x, \theta) q_f(z|x, \phi)} \bigg{]} \\
&=   \ln \ {p(x|z, \theta)} +  \ln  \frac{p(z|\theta)}{q_f(z|x, \phi)} +  \ln  \frac{q_f(z|x, \phi)}{q(z|x, \theta)} \\ \label{eq:logp_num_1}
&= \ln \ {p(x|z, \theta)} -  \ln  \frac{q_f(z|x, \phi)}{p(z|\theta)} +  \ln  \frac{q_f(z|x, \phi)}{q(z|x, \theta)} \ . 
\end{align}
We now take the expectation of (\ref{eq:logp_num_1}) with respect to $q_f(z|x,\phi)$:
\begin{align}\label{eq:lhs_logp_expectation}
\mathbb{E}_{z|x,\phi} \big{[} \ln \ p(x|\theta) \big{]} &= \int_z q_f(z|x,\phi) \ \ln \ p(x|\theta) \ dz \\
&= \ln \ p(x|\theta)  \int_z q_f(z|x,\phi)  \ dz \\
& =  \ln \ p(x|\theta) 
\end{align}
because $q_f(z|x,\phi)$ is a probability density function and its integral over the domain of $z$ equals 1.
The expectation of the term on the right-hand side of (\ref{eq:logp_num_1})  with respect $q_f(z|x,\phi)$ is 
\begin{multline}\label{eq:rhs_logp_expectation}
\mathbb{E}_{z|x,\phi} \bigg{[} \ln \ {p(x|z, \theta)} -  \ln  \frac{q_f(z|x, \phi)}{p(z|\theta)} +  \ln  \frac{q_f(z|x, \phi)}{q(z|x, \theta)} \bigg{]} \\=
\mathbb{E}_{z|x,\phi}  \big{[} \ln \ {p(x|z, \theta)} \big{]} - D_\mathtt{KL}\big{(}{q_f(z|x,\phi)} || {p(z|\theta)}\big{)} \\+  D_\mathtt{KL}\big{(}{q_f(z|x,\phi)} || {q(z|x,\theta)}\big{)}
\end{multline}
where
\begin{align}
D_\mathtt{KL}\big{(}{q_f(z|x,\phi)} || {p(z|\theta)}\big{)} &=  \mathbb{E}_{z|x,\phi}  \bigg{[} \ln  \frac{q_f(z|x, \phi)}{p(z|\theta)}  \bigg{]} 
\end{align}
and
\begin{align}
D_\mathtt{KL}\big{(}{q_f(z|x,\phi)} || {q(z|x,\theta)}\big{)} &=  \mathbb{E}_{z|x,\phi}  \bigg{[}  \ln \frac{q_f(z|x, \phi)}{q(z|x, \theta)}  \bigg{]} \ .
\end{align}

Combining (\ref{eq:logp_num_1}), (\ref{eq:lhs_logp_expectation}), and (\ref{eq:rhs_logp_expectation}) gives the following:
\begin{multline}
\ln p(x|\theta) = \underbrace{\mathbb{E}_{z|x,\phi}\big{[} \ln p(x|z, \theta) \big{]} }_{\substack{\text{\sf \footnotesize \textcolor{blue}{Decoding}}}}  - \underbrace{D_\mathtt{KL}\big{(}{q_f(z|x,\phi)} || {p(z|\theta)}\big{)}}_{\substack{\text{\sf \footnotesize \textcolor{red}{Encoding}}}}   \\+ \underbrace{D_\mathtt{KL}\big{(}{q_f(z|x,\phi)} || {q(z|x,\theta)}\big{)}}_{\substack{\text{\sf \footnotesize \textcolor[rgb]{0.0, 0.4,0.0}{Error-gap}}}}  .
\end{multline}
The KL-divergence between ${q_f(z|x,\phi)}$ and $q(z|x,\theta)$ yields the following inequality because of Jensen's inequality: 
\begin{align}
D_\mathtt{KL}\big{(}{q_f(z|x,\phi)} & || {q(z|x,\theta)}\big{)} =   \mathbb{E}_{z|x,\phi}  \bigg{[}  \ln \frac{q_f(z|x, \phi)}{q(z|x, \theta)}  \bigg{]} \\
&= \int_z  q_f(z|x, \phi)  \ln \frac{q_f(z|x, \phi)}{q(z|x, \theta)} \ dz \\
&= -\int_z  q_f(z|x, \phi)  \ln \frac{q(z|x, \theta)}{q_f(z|x, \phi)} \\
& \geq - \ln \int_z  \cancel{q_f(z|x, \phi)}  \frac{{q(z|x, \theta)}}{\cancel{q_f(z|x, \phi)}} \\
& = -\ln \int_z  \ {q(z|x, \theta)} \ dz \\
&= - \ln \ 1  = 0
\end{align}
because the negative of the natural logarithm is convex.
So we have
\begin{align}
\ln p(x|\theta)  &\geq  \underbrace{\mathbb{E}_{z|x,\phi}\big{[} \ln p(x|z, \theta) \big{]}   - D_\mathtt{KL}\big{(}{q_f(z|x,\phi)} || {p(z|\theta)}\big{)}}_{\substack{\text{\sf \small \textcolor{Brown}{$\mathcal{L}(x,\theta, \phi)$}}}} 
\end{align}
and 
\begin{align}\label{eq:elbo_vae}
\mathcal{L}(x,\theta, \phi) &= \mathbb{E}_{z|x,\phi}\big{[} \ln p(x|z, \theta) \big{]}   - D_\mathtt{KL}\big{(}{q_f(z|x,\phi)} || {p(z|\theta)}\big{)}
\end{align}
where $\mathcal{L}(x,\theta, \phi)$ is the evidence lower bound (ELBO) on the data log-likelihood $p(x|\theta)$.\\

Unidirectional VAEs train on the estimate $\tilde{\mathcal L}_{\mathtt{ELBO}}(x,\theta,\phi)$ of the ELBO using the ordinary or unidirectional backpropagation (BP).
This estimate involves using the forward pass $q_f(z|x,\phi)$ to approximate the intractable encoding model $q(z|x,\theta)$ and the forward pass $p_f(x|z,\theta)$ to approximate the encoding model.
The update rules for the encoder and decoder networks at the $(n + 1)^{th}$ iteration or training epoch are
\begin{align}
\theta^{(n+1)} &=\theta^{(n)} + \eta \nabla_\theta  \ \tilde{\mathcal{L}}_{\mathtt{ELBO}}(x,\theta,\phi) \Bigg{|}_{\theta = \theta^{(n)}, \phi = \phi^{(n)}} \\
\phi^{(n+1)} &=\phi^{(n)} + \eta \nabla_\phi \  \tilde{\mathcal{L}}_{\mathtt{ELBO}}(x,\theta,\phi) \Bigg{|}_{\theta = \theta^{(n)}, \phi = \phi^{(n)}}
\end{align}
where $\eta$ is the learning rate, $\phi^{(n)}$ is the encoder parameter, and $\theta^{(n)}$ is the decoder parameter after  $n$ training iterations.

\section{Bidirectional Variational Autoencoders: BELBO Maximization}
\label{sec:bvae}


Bidirectional VAEs use the directional likelihoods of a network with parameter $\theta$ to approximate the data log-likelihood $\ln p(x|\theta)$.  
They use the same bidirectional associative network to model the encoding and decoding phases. 
The forward-pass likelihood $q_f(z|x,\theta)$ models the encoding and the backward-pass likelihood $p_b(x|z,\theta)$ models the decoding.
So BVAEs do not need an extra likelihood $q(z|x,\phi)$ or an extra network with parameter $\phi$. 
We call this BELBO or bidirectional ELBO maximization.

The data likelihood is
\begin{align}
p(x|\theta) &= \frac{p(x|z, \theta) p(z|\theta)}{q(z|x, \theta)}
 .
\end{align}
Then we have
\begin{align} 
\ln p(x|\theta) &= \ln  \bigg{[}\frac{p(x|z, \theta) p(z|\theta)}{q(z|x, \theta)} \bigg{]} \\
&= \ln  \bigg{[}\frac{p(x|z, \theta) p(z|\theta) q_f(z|x, \theta) }{q(z|x, \theta) q_f(z|x, \theta) } \bigg{]} \\ \nonumber
&= \ln  \big{[} p(x|z, \theta)\big{]} -  \ln \bigg{[}\frac{q_f(z|x,\theta)}{p(z|\theta)} \bigg{]} \\ \label{eq:logp_bvae_def}
& \hspace{0.8in} +  \ln \bigg{[}\frac{q_f(z|x,\theta)}{q(z|x, \theta)} \bigg{]} . 
\end{align}
We now take the expectation of (\ref{eq:logp_bvae_def}) with respect to $q_f(z|x,\theta)$.
Let us consider the left-hand side of (\ref{eq:logp_bvae_def}).
We have
\begin{align}
\mathbb{E}_{z|x,\theta}\big{[}\ln p(x|\theta) \big{]} &= \int_z \ q_f(z|x,\theta) \ln  p(x|\theta)  \ dz \\
&= \ln p(x|\theta) \int_z  \ q_f(z|x,\theta) \ dz \\ 
&= \ln p(x|\theta) .
\end{align}
The expectation of the right-hand term is
\begin{multline}
\mathbb{E}_{z|x,\theta} \bigg{[} \ln \ {p(x|z, \theta)} -  \ln  \frac{q_f(z|x, \theta)}{p(z|\theta)}  + \ln \frac{q_f(z|x,\theta)}{q(z|x, \theta)}\bigg{]} \\=
\mathbb{E}_{z|x,\theta}  \big{[} \ln \ {p(x|z, \theta)} \big{]} - D_\mathtt{KL}\big{(}{q_f(z|x,\theta)} || {p(z|\theta)}\big{)} \\ +  D_\mathtt{KL}\big{(}{q_f(z|x,\theta)} || {q(z|x,\theta)}\big{)}
\end{multline}
where
\begin{align}
 \mathbb{E}_{z|x,\theta}  \bigg{[} \ln \frac{q_f(z|x, \theta)}{p(z|\theta)}  \bigg{]} &= D_\mathtt{KL}\big{(}{q_f(z|x,\theta)} || {p(z|\theta)}\big{)} 
 \end{align}
 and
 \begin{align}
 \mathbb{E}_{z|x,\theta}  \bigg{[} \ln \frac{q_f(z|x, \theta)}{q(z|x,\theta)}  \bigg{]}  &= D_\mathtt{KL}\big{(}{q_f(z|x,\theta)} || {q(z|x,\theta)}\big{)}  \ .
\end{align}

The corresponding data log-likelihood of a BVAE with  parameter $\theta$ is
\begin{multline}\label{eq:logp_x_z}
\ln p(x|\theta) = \underbrace{\mathbb{E}_{z|x,\theta}\big{[} \ln p(x|z, \theta) \big{]} }_{\substack{\text{\sf \footnotesize \textcolor{blue}{Decoding}}}}  - \underbrace{D_\mathtt{KL}\big{(}{q_f(z|x,\theta)} || {p(z|\theta)}\big{)}}_{\substack{\text{\sf \footnotesize \textcolor{red}{Encoding}}}}   \\+ \underbrace{D_\mathtt{KL}\big{(}{q_f(z|x,\theta)} || {q(z|x,\theta)}\big{)}}_{\substack{\text{\sf \footnotesize \textcolor[rgb]{0.0, 0.4,0.0}{Error-gap}}}}  .
\end{multline} 
The log-likelihood of the BVAE is  such that
\begin{align}
\ln p(x|\theta)  &\geq  \underbrace{\mathbb{E}_{z|x,\theta}\big{[} \ln p(x|z, \theta) \big{]}   - D_\mathtt{KL}\big{(}{q_f(z|x,\theta)} || {p(z|\theta)}\big{)}}_{\substack{\text{\sf \small \textcolor{Brown}{$\mathcal{L}_{\mathtt{BELBO}}(x,\theta)$}}}} 
\end{align}
because the KL-divergence  $D_\mathtt{KL}\big{(}{q_f(z|x,\theta)} || {q(z|x,\theta)}\big{)} \geq 0$. 
So we have
\begin{align}\label{eq:elbo_bvae}
\mathcal{L}_{\mathtt{BELBO}}(x,\theta) &= \mathbb{E}_{z|x,\theta}\big{[} \ln p(x|z, \theta) \big{]}   - D_\mathtt{KL}\big{(}{q_f(z|x,\theta)} || {p(z|\theta)}\big{)}
\end{align}
where $\mathcal{L}_{\mathtt{BELBO}}(x,\theta)$ is the BELBO on $\ln p(x|\theta)$ and the expectation $\mathbb{E}_{z|x,\theta}$ is with respect to $q_f(z|x,\theta)$. \\


Bidirectional VAEs train on the estimate ${\tilde{\mathcal L}}_{\mathtt{BELBO}}(x,\theta)$ of the BELBO that uses bidirectional neural representation \cite{AdigunK20_v1}.
This estimate uses the forward pass $q_f(z|x,\theta)$ to approximate the intractable encoding model $q(z|x,\theta)$ and uses the reverse pass $p_b(x|z,\theta)$ to approximate the decoding model.
The update rule at the $(n + 1)^{th}$ iteration or training epoch is
\begin{align}
\theta^{(n+1)} &=\theta^{(n)} + \eta \nabla_\theta {\tilde{\mathcal{L}}}_{\mathtt{BELBO}}(x, \theta) \Bigg{|}_{\theta = \theta^{(n)}}
\end{align}
where $\eta$ is the learning rate and $\theta^{(n)}$ is the autoencoder network parameter just after the $n^{th}$ training iteration.
Figure \ref{fig:bidirectional_BP_training} shows the probabilistic approximation of a BVAE with the directional likelihoods of a bidirectional network.\\

\begin{figure*}[t]
  \centering
   \begin{subfigure}[b]{0.53\linewidth}
         \centering
         \includegraphics[width=\textwidth, height=0.68\textwidth]{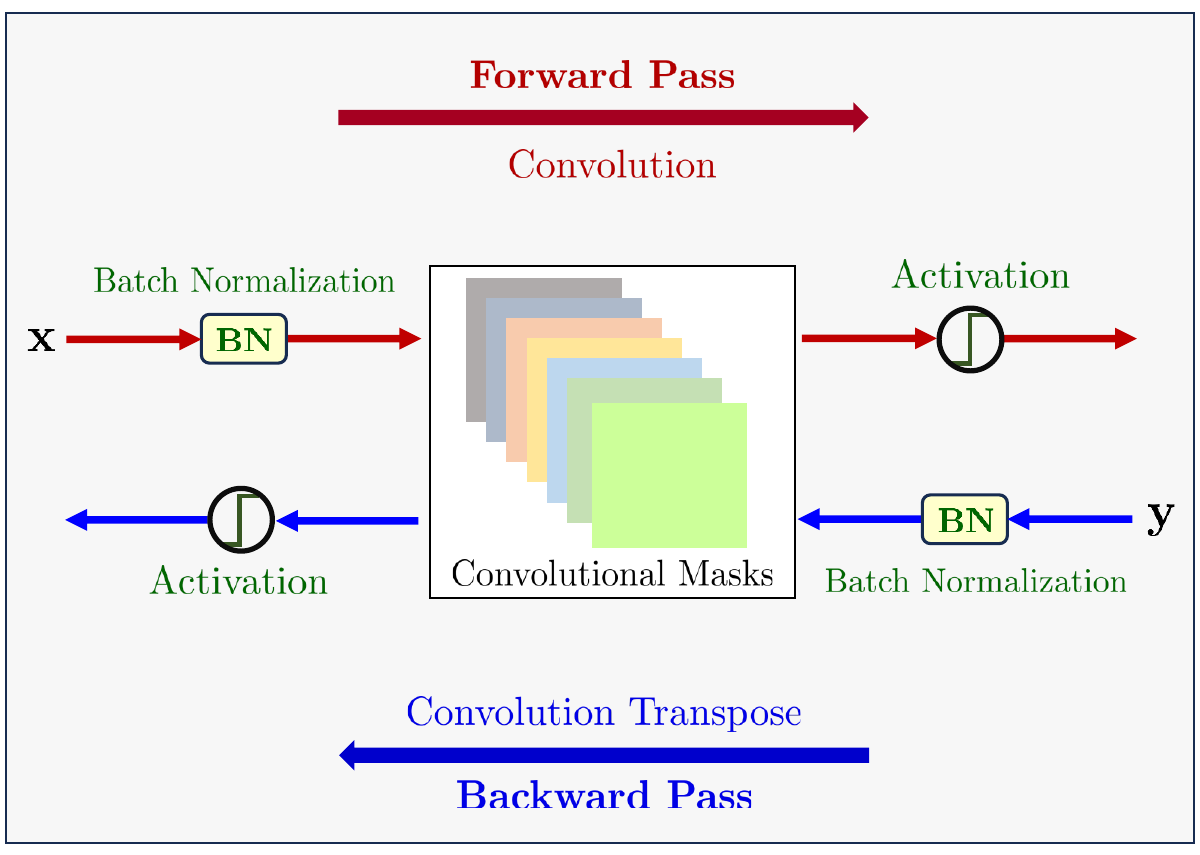}
         \caption{Bidirectional convolutional (Bi-Conv2D) layer}
         \label{fig:bi_convd2d}
   \end{subfigure}
   \hfill
   \begin{subfigure}[b]{0.33\linewidth}
         \centering
         \includegraphics[width=\textwidth, height=1.1\textwidth]{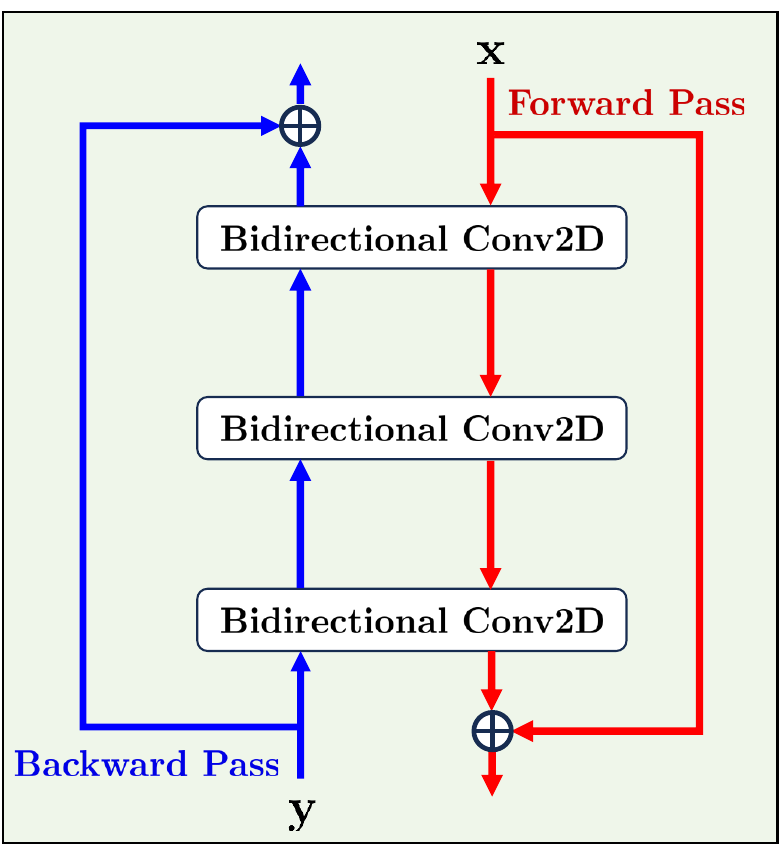}
         \caption{Bidirectional residual block (Bi-ResBlock) }
         \label{fig:bi_resblock}
   \end{subfigure}
      \hfill
   \begin{subfigure}[b]{0.80\linewidth}
         \centering
         \vspace{0.3in}
         \includegraphics[width=\textwidth]{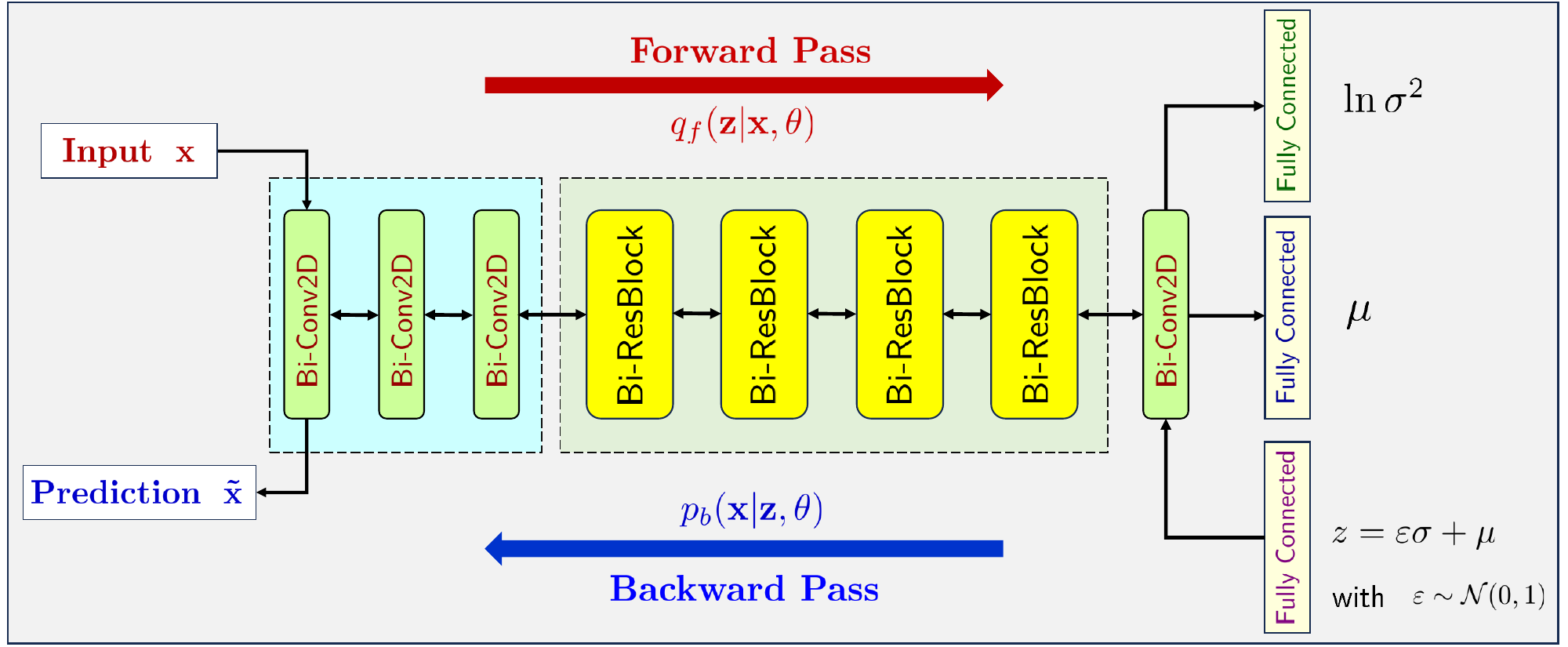}
         \caption{Bidirectional residual VAE network}
         \label{fig:bi_resnet}
   \end{subfigure}
   \caption{\small Bidirectional VAE with residual network architecture:  
This cuts the tunable parameters roughly in half compared with unidirectional VAEs.
(a) is the bidirectional convolutional layer.
Convolution runs in the forward pass and convolution transpose runs in reverse with the same set of convolution masks.
(b) is the architecture of a bidirectional residual block with bidirectional skip connections.}
   \label{fig:bidirectional_layers}
\end{figure*}

\begin{algorithm}[!t]
 \caption{\hspace{-0.05in}:  \small BVAE Training  with bidirectional backpropagation}
 \begin{algorithmic}[1]
 \renewcommand{\algorithmicrequire}{\textbf{Input:}}
 \renewcommand{\algorithmicensure}{\textbf{Output:}}
 \REQUIRE \small{Data $\{ {\bf x}_n \}_{n=1}^{N}$ and latent space dimension $J$.}\\
\small{\textit{Initialize}: Synaptic weights $\theta \in \{ {\mathbf N}_\theta,  {\mathbf V}_{\theta}, {\mathbf W}_{\theta}\}$ , learning rate $\alpha$, and other hyper-parameters.}
   \FOR {\small{iteration $t = 1,2, .....$}}
  \STATE {\small Pick a mini-batch $\{ {\bf x}_m \}_{m=1}^{B}$  of $B$ samples}
   \FOR {$m = 1,2, ...., B$}
 \STATE \text{\small \textcolor{blue}{Forward Pass (Encoding):}} Predict the variational mean and log-covariance:
 \[
 {{{\hat{\bf{\mu}}}}_m} = {\bf W}_{\theta} \Big{(} {\mathbf N}_{\theta}  \big{(} {\bf x}_m \big{)} \Big{)} \]  and \[
 \ln{{{\hat{\sigma}}}_m^2} = {\bf V}_{\theta} \Big{(} {\mathbf N}_{\theta}  \big{(} {\bf x}_m \big{)} \Big{)} \]
 \STATE {\small Sample the latent features ${\bf z}_m$ from the variational Gaussian distribution with condition ${\bf x}_m$:}
 \[ {\bf \varepsilon}_m \sim \mathcal{N}({\bf 0}, {\bf I})  \ \ \ \ \text{and} \ \ \ \ {\bf z}_{m} = {{\hat{\mu}}}_{m} + {\bf \varepsilon}_m  \cdot  {{\hat{\sigma}}}_m  \]
  \STATE \text{\small \textcolor{red}{Backward Pass (Decoding):}} Map the latent variable back to the input space:
   \[
 {{{\hat{a}}}_{m}^{(x)}} = {{\mathbf N}_{\theta}}^{\mathsf{T}} \Big{(} {{\bf W}_{\theta}}^{\mathsf{T}}  \big{(} {\bf z}_m \big{)} \Big{)} \]
  \ENDFOR
  \STATE Estimate the negative log-likelihood $\mathtt{NLL({\mathbf{x},\theta})}$:
  \[ \mathtt{KLD}({\bf x}, \theta) =  \frac{1}{B} \sum_{m=1}^{B} D_{\mathtt{KL}} \big{(} \mathcal{N}(\hat{\mu}_m, {\bf Diag}(\hat{\sigma}^2_m)) || \mathcal{N}(\bf{0}, \bf{I} ) \big{)} \]
  \[ \mathtt{BCE}({\bf x}, \theta) = - \frac{1}{B} \sum_{m=1}^{B} \ln p_b({\bf x}_m|{\bf z}_m, \theta) \hspace{0.95in}\]
  \[ \tilde{\mathcal{L}}_{\mathtt{ELBO}}({\bf x}, \theta)  = \mathtt{BCE}({\bf x}, \theta) \ + \ \mathtt{KLD}({\bf x}, \theta)  \hspace{1.20in}\]
  \STATE Update $\theta$ by backpropagating  $\tilde{\mathcal{L}}_{\mathtt{ELBO}}({\bf x}, \theta)$ through the weights.
   \ENDFOR
 \RETURN $\theta$ 
 \end{algorithmic} 
 \end{algorithm}

\section{Simulations}
\label{sec:exp_setup}
 
 We compared the performance of unidirectional VAEs and bidirectional VAEs using different tasks, datasets, network architectures, and loss functions. 
We first describe the image test sets for our experiments.
 
 \subsection{Datasets}
 The simulations compared results on four standard image datasets: MNIST handwritten digits \cite{lecun1998mnist}, Fashion-MNIST \cite{xiao2017fashion}, CIFAR-10 \cite{krizhevsky2009learning}, and CelebA \cite{liu2015faceattributes} datasets.

 The MNIST handwritten digit dataset contains 10 classes of handwritten digits $\{ 0,1,2,3,4,5,6,7,8,9\}$. 
 This dataset consists of 60,000 training samples with 6,000 samples per class, and 10,000 test samples with 1,000 samples per class
 Each image is a single-channel image with dimension $28 \times 28$.
 
 The Fashion-MNIST dataset is a database of fashion images.
 It is made of 10 classes namely  ankle boot, bag, coat, dress, pullover, sandal, shirt, sneaker, trouser, and t-shirt / top.
 Each class has 6,000 training samples and 1,000 testing samples.
 Each image is also a single-channel image with dimension $28\times 28$.
 
The CIFAR-10 dataset consists of 60,000 color images  from 10 categories.
Each image has size $32 \times 32 \times 3$.
The 10 pattern categories are airplane, automobile, bird, cat, deer, dog, frog, horse, ship, and truck.
Each class consists of 5,000 training samples and 1,000 testing samples.

The CelebA  dataset is a large-scale face dataset of 10,177 celebrities \cite{liu2015faceattributes}.
This dataset is made up of 202,599 color (three-channel) images.
This is not a balanced dataset.
The number of images per celebrity varies between $1-35$.
We divided the dataset into two splits of 9,160 celebrities for training and 1,017 celebrities for testing the VAEs.
This resulted in 185,133 training samples and 17,466 testing samples. 
We resized each image to $64 \times 64 \times3$.

\begin{figure}[tbp]
  \centering
   \begin{subfigure}[b]{0.65\linewidth}
         \centering
         \includegraphics[width=\textwidth]{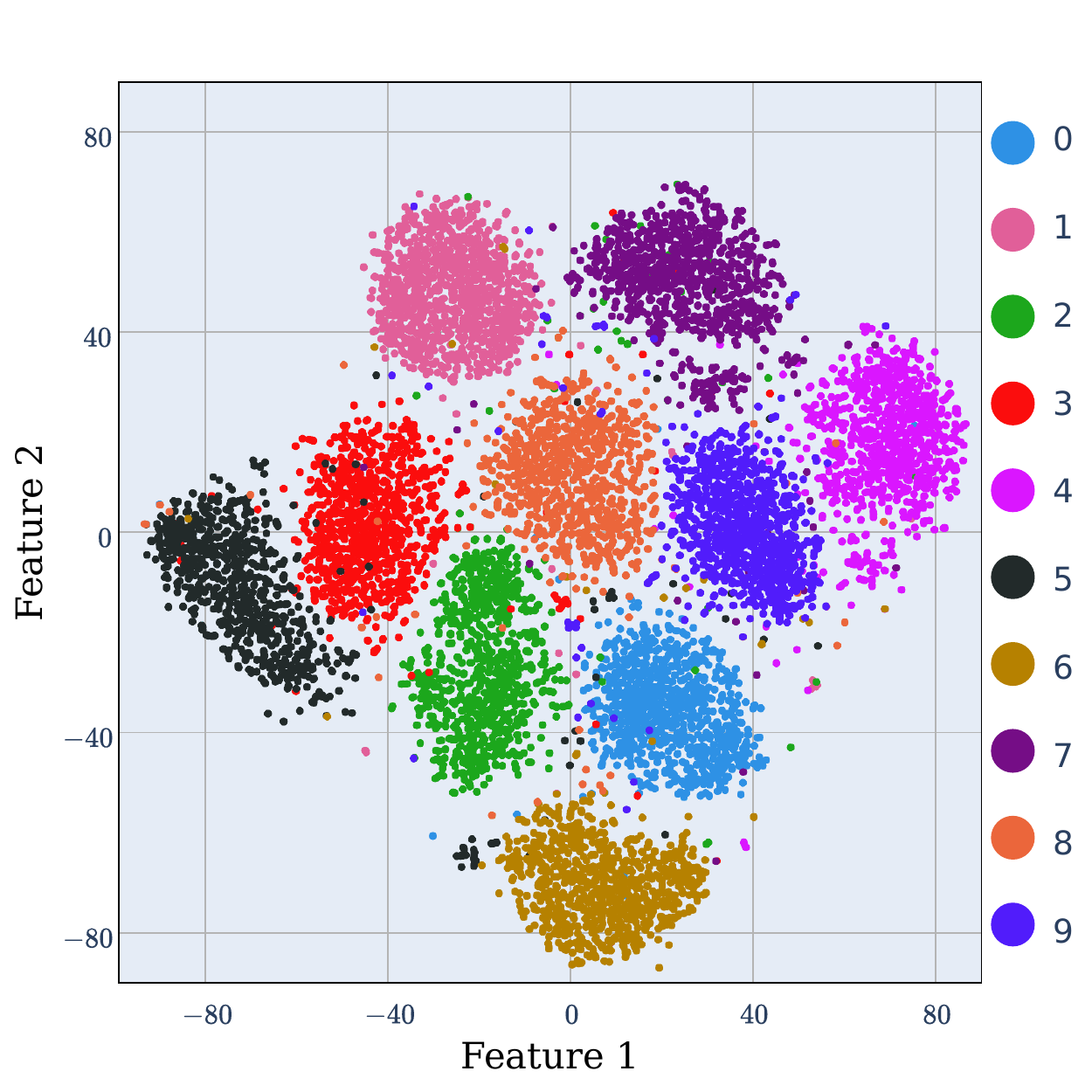}
         \caption{Unidirectional $\beta$-VAE }
         \label{fig:alpha_1.0_beta_2.0}
   \end{subfigure}
   \hfill
   \begin{subfigure}[b]{0.65\linewidth}
         \centering
         \includegraphics[width=\textwidth]{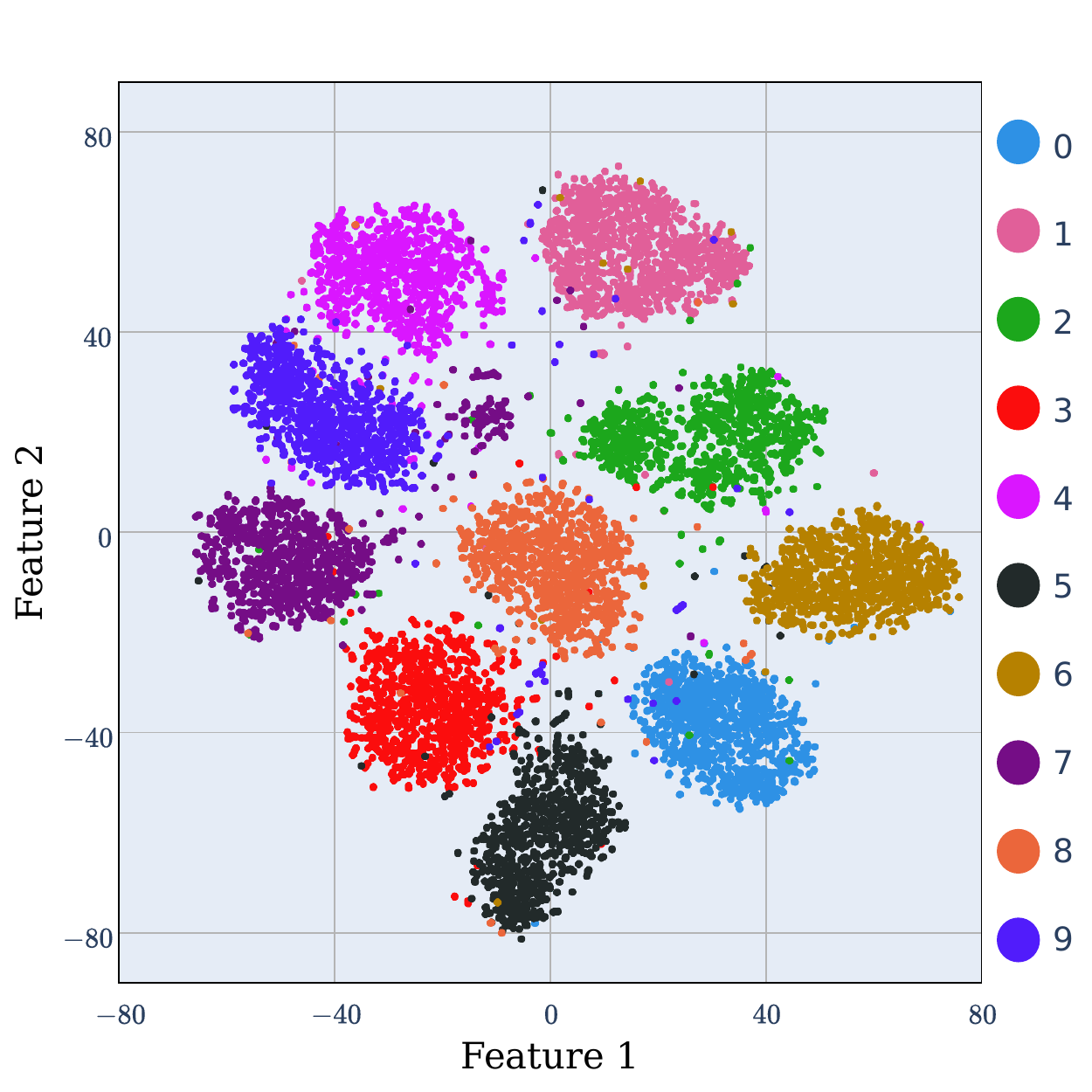}
         \caption{Bidirectional $\beta$-VAE}
         \label{fig:alpha_1.0_beta_2.0}
   \end{subfigure}
   \caption{\small{$t$-SNE embedding for the MNIST handwritten digit dataset: Latent space dimension is $128$.
   (a) A simple linear classifier that trained on the unidirectional VAE-compressed features achieved a $95.2\%$ accuracy.
   (b) The simple classifier achieved $97.32\%$ accuracy when it trained on the BVAE-compressed features.}}
   \label{fig:tsne}
\end{figure}

 \subsection{Tasks}

We compared the performance of bidirectional VAEs and unidirectional VAEs on the following four tasks.
 
\subsubsection{Image compression and reconstruction}


We explore the self-mapping of the image datasets with unidirectional and bidirectional VAEs.
This involves the encoding of images with latent variable $z$ and the subsequent decoding to reconstruct the image after the latent sampling.
We evaluated the performance of VAEs on this task using the Peak Signal-to-Noise Ratio (PSNR), Fr\'echet Inception Distance (FID) \cite{HeuselRUNH17, frechet1957distance}, and Structural Similarity Index Measure (SSIM) \cite{WangBSS04}.

\begin{table*}[!t]
  \centering
  \caption{\small MNIST handwritten digits dataset with VAEs. We used the residual network architecture. The dimension of the latent variable is $64$. 
  The BVAEs each used $42.2$MB storage memory and the VAEs each used $84.4$MB storage memory.}
  \begin{tabular}{|l|c|ccc|cccc|}
    \hline
    \multicolumn{1}{|l|}{\multirow{3}{*}{\textbf{Model}}} & \multirow{3}{*}{\bf Parameters} & \multicolumn{3}{c|}{\multirow{2}{*}{\bf Generative Task}} & \multicolumn{4}{c|}{\multirow{2}{*}{ {\bf Reconstruction} and {\bf Classification}}} \\
    \multicolumn{1}{|l|}{\multirow{1}{*}}  &{} & \multicolumn{3}{|c|}{\multirow{2}{*}{}} & \multicolumn{4}{c|}{\multirow{2}{*}{}}\\
    {} & {}& {\bf NLL} $\downarrow$ & {\bf AU}$$ & {\bf FID}$\downarrow$  &  {\bf PSNR} $\uparrow$ & {\bf SSIM}$\uparrow$ &  {\bf rFID}$\downarrow$  & {\bf Accuracy} $\uparrow$ \\
    \hline
     VAE  & $22.1$M & ${86.72}$& {$18$} &{$4.340$}  & ${20.59}$ & ${ 0.8959}$& ${3.510}$ & {$94.97$\%} \\
    \textcolor[rgb]{0.,0.,0.9}{{Bidirectional} VAE} & \textcolor[rgb]{0,0,0.9}{{$11.1$M}}  & \textcolor[rgb]{0,0,0.9}{$86.69$}& \textcolor[rgb]{0,0,0.9}{\bf{$15$}}& \textcolor[rgb]{0,0,0.9}{$3.370$} &  \textcolor[rgb]{0,0,0.9}{$20.64$} & \textcolor[rgb]{0,0,0.9}{$0.8988$}& \textcolor[rgb]{0,0,0.9}{$3.050$} & \textcolor[rgb]{0,0,0.9}{\bf{$95.26\%$}} \\
        \hline
        \hline
   $\beta$-VAE  $(\beta=0.5)$ & $22.1$M & ${88.12}$& {$23$} &{$2.021$}  & ${22.19}$ & ${ 0.9306}$& ${1.711}$ & {$93.15$\%} \\
    \textcolor[rgb]{0.,0.,0.9}{{Bidirectional} $\beta$-VAE $(\beta=0.5)$} & \textcolor[rgb]{0,0,0.9}{{$11.1$M}}  & \textcolor[rgb]{0,0,0.9}{$88.09$}& \textcolor[rgb]{0,0,0.9}{\bf{$26$}}& \textcolor[rgb]{0,0,0.9}{$2.034$} &  \textcolor[rgb]{0,0,0.9}{$22.39$} & \textcolor[rgb]{0,0,0.9}{$0.9332$}& \textcolor[rgb]{0,0,0.9}{$1.702$} & \textcolor[rgb]{0,0,0.9}{\bf{$94.32\%$}} \\
    \hline
       $\beta$-VAE  $(\beta=1.5)$ & $22.1$M & ${86.99}$& {$15$} &{$4.855$}  & ${19.75}$ & ${ 0.8726}$& ${4.370}$ & {$95.20$\%} \\
    \textcolor[rgb]{0.,0.,0.9}{{Bidirectional} $\beta$-VAE $(\beta=1.5)$} & \textcolor[rgb]{0,0,0.9}{{$11.1$M}}  & \textcolor[rgb]{0,0,0.9}{$87.21$}& \textcolor[rgb]{0,0,0.9}{\bf{$15$}}& \textcolor[rgb]{0,0,0.9}{$5.589$} &  \textcolor[rgb]{0,0,0.9}{$19.80$} & \textcolor[rgb]{0,0,0.9}{$0.8736$}& \textcolor[rgb]{0,0,0.9}{$5.011$} & \textcolor[rgb]{0,0,0.9}{\bf{$96.19\%$}} \\
    \hline
            \hline
       $\beta$-TCVAE  $(\beta=0.5)$ & $22.1$M & ${88.18}$& {$24$} &{$1.975$}  & ${22.21}$ & ${ 0.9299}$& ${1.662}$ & {$93.37\%$} \\
    \textcolor[rgb]{0.,0.,0.9}{{Bidirectional} $\beta$-TCVAE  $(\beta=0.5)$} & \textcolor[rgb]{0,0,0.9}{{$11.1$M}}  & \textcolor[rgb]{0,0,0.9}{$87.99$}& \textcolor[rgb]{0,0,0.9}{\bf{$26$}}& \textcolor[rgb]{0,0,0.9}{$2.000$} &  \textcolor[rgb]{0,0,0.9}{$22.30$} & \textcolor[rgb]{0,0,0.9}{$0.9325$}& \textcolor[rgb]{0,0,0.9}{$1.760$} & \textcolor[rgb]{0,0,0.9}{\bf{$94.17\%$}} \\
        \hline
       $\beta$-TCVAE  $(\beta=1.0)$ & $22.1$M & ${86.87}$& {$18$} &{$3.544$}  & ${20.68}$ & ${0.8988}$& ${3.143}$ & {$95.08\%$} \\
    \textcolor[rgb]{0.,0.,0.9}{{Bidirectional} $\beta$-TCVAE $(\beta=1.0)$} & \textcolor[rgb]{0,0,0.9}{{$11.1$M}}  & \textcolor[rgb]{0,0,0.9}{$86.68$}& \textcolor[rgb]{0,0,0.9}{\bf{$18$}}& \textcolor[rgb]{0,0,0.9}{$3.419$} &  \textcolor[rgb]{0,0,0.9}{$20.75$} & \textcolor[rgb]{0,0,0.9}{$0.9007$}& \textcolor[rgb]{0,0,0.9}{$3.085$} & \textcolor[rgb]{0,0,0.9}{\bf{$95.66\%$}} \\
        \hline
      $\beta$-TCVAE  $(\beta=1.5)$ & $22.1$M & ${87.20}$& {$14$} &{$5.623$}  & ${19.64}$ & ${ 0.8690}$& ${5.052}$ & {$95.56\%$} \\
    \textcolor[rgb]{0.,0.,0.9}{{Bidirectional $\beta$-TCVAE} $(\beta=1.5)$} & \textcolor[rgb]{0,0,0.9}{{$11.1$M}}  & \textcolor[rgb]{0,0,0.9}{$86.90$}& \textcolor[rgb]{0,0,0.9}{\bf{$15$}}& \textcolor[rgb]{0,0,0.9}{$4.968$} &  \textcolor[rgb]{0,0,0.9}{$19.86$} & \textcolor[rgb]{0,0,0.9}{$0.8756$}& \textcolor[rgb]{0,0,0.9}{$4.548$} & \textcolor[rgb]{0,0,0.9}{\bf{$96.54\%$}} \\
        \hline
        \hline
              IWAE  & $22.1$M & ${86.02}$& {$20$} &{$2.958$}  & ${19.30}$ & ${0.8627}$& ${2.774}$ & {$96.40\%$} \\
    \textcolor[rgb]{0.,0.,0.9}{{Bidirectional} IWAE}  & \textcolor[rgb]{0,0,0.9}{{$11.1$M}}  & \textcolor[rgb]{0,0,0.9}{$86.00$}& \textcolor[rgb]{0,0,0.9}{\bf{$19$}}& \textcolor[rgb]{0,0,0.9}{$3.330$} &  \textcolor[rgb]{0,0,0.9}{$19.21$} & \textcolor[rgb]{0,0,0.9}{$0.8587$}& \textcolor[rgb]{0,0,0.9}{$3.030$} & \textcolor[rgb]{0,0,0.9}{\bf{$96.29\%$}} \\
        \hline
  \end{tabular}
  \label{tab:mnist_latent_dim_64}
\end{table*}

\subsubsection{Downstream image classification} 
We trained simple classifiers on the latent space features of VAEs.
These VAEs compress the input images and a simple classifier maps the latent features to their corresponding classes. 
We evaluated the classification accuracy of this downstream classification.

The VAE-extracted features for the MNIST handwritten dataset trained on simple linear classifiers.
We used a neural classifier with one hidden layer of $256$ logistic hidden neurons to classify the VAE-extracted features from the Fashion-MNIST and CIFAR-10 datasets.

We used the $t$-distributed stochastic neighbor embedding ($t$-SNE) method to visualize the reduced features.
This method uses a statistical approach to map the high-dimensional representation of data $\big{\{}{\bf x}_i\big{\}}_{i=1}^{N}$ to their respective low-dimensional representation $\big{\{}{\bf y}_i\big{\}}_{i=1}^{N}$ based on the similarity of the datapoints \cite{van2008visualizing}. 
This low-dimensional representation provides insight into the degree of separability among the classes.

\subsubsection{Image generation}

We compared the generative performance of bidirectional VAEs with their corresponding unidirectional ones.
These VAEs trained with the Gaussian latent distribution  $\mathcal{N}(\bf{0}, \bf{I})$.
We tested the VAEs using both Gaussian sampler and Gaussian mixture model samplers post-training \cite{GhoshSVBS20}.
We used the estimates of the negative of the data log-likelihood (NLL) and the number of active latent units (AU) \cite{BurdaGS15} as the quantitative metrics for performance evaluation.

\subsubsection{Image interpolation}

We conducted linear interpolation of samples. 
The interpolations involve a convex combination of two images over 10 steps. 
The encoding step transforms the mixture of two images in the latent space. 
The decoding step reconstructs the interpolated samples.

\

\subsection{Model Architecture}

We used different neural network architectures for various datasets and tasks.

\textit{Variational Autoencoders}: We used deep convolutional and residual neural network architectures.
VAEs that trained on the MNIST handwritten and Fashion-MNIST used the residual architecture.
Figure \ref{fig:bi_resnet} shows the architecture of the bidirectional residual networks that trained on the MNIST datasets. 
The unidirectional VAEs with this architecture used two such networks each: one for encoding and the other for decoding.
The BVAEs used just one of such network each. 
Encoding runs in the forward pass and decoding runs in the backward pass. 

Each of the encoder and decoder networks that trained on the CIFAR-10 dataset used six convolutional layers and two fully connected layers.
The corresponding BVAEs used only one network for encoding and decoding each.  
The dimension of the hidden convolutional layers  is $ \{64  \leftrightarrow 128  \leftrightarrow   256  \leftrightarrow  512 \leftrightarrow  1024  \leftrightarrow 2048 \} $. 
The dimension of the fully connected layers is $\{2048 \leftrightarrow 1024 \leftrightarrow 64 \}$.

The configuration of the VAEs that trained on the CelebA dataset differs slightly.
The sub-networks each used  nine convolutional layers and two fully connected layers.
The dimension of the hidden convolutional layers  is $ \{128  \leftrightarrow 128  \leftrightarrow   192 \leftrightarrow  256 \leftrightarrow  384  \leftrightarrow 512  \leftrightarrow 768  \leftrightarrow 1024  \leftrightarrow 1024 \} $. 
The dimension of the fully connected layers is $\{4096 \leftrightarrow 2048 \leftrightarrow 256 \}$.



The VAEs used generalized nonvanishing (G-NoVa) hidden neurons \cite{AdigunK22_icmla22, adigun2021deeper}.
The G-NoVa activation $a(x)$ of input $x$ is
\begin{align}
a(x) &= \alpha x + x \sigma(\beta x)  = \alpha x + \frac{x}{1+e^{-\beta x}} 
\end{align}
 where $\alpha > 0$ and $\beta > 0$.
Each layer of a BVAE performs probabilistic inference in both the forward and backward passes.
The convolutional layers use bidirectional kernels.
The kernels run convolution in the forward pass and transposed convolution in the backward pass. 
Transposed convolution projects feature maps to a higher-dimensional space \cite{dumoulin2016guide}.

\begin{figure*}[t]
   \begin{subfigure}[b]{0.24\linewidth}
         \centering
         \includegraphics[width=\textwidth]{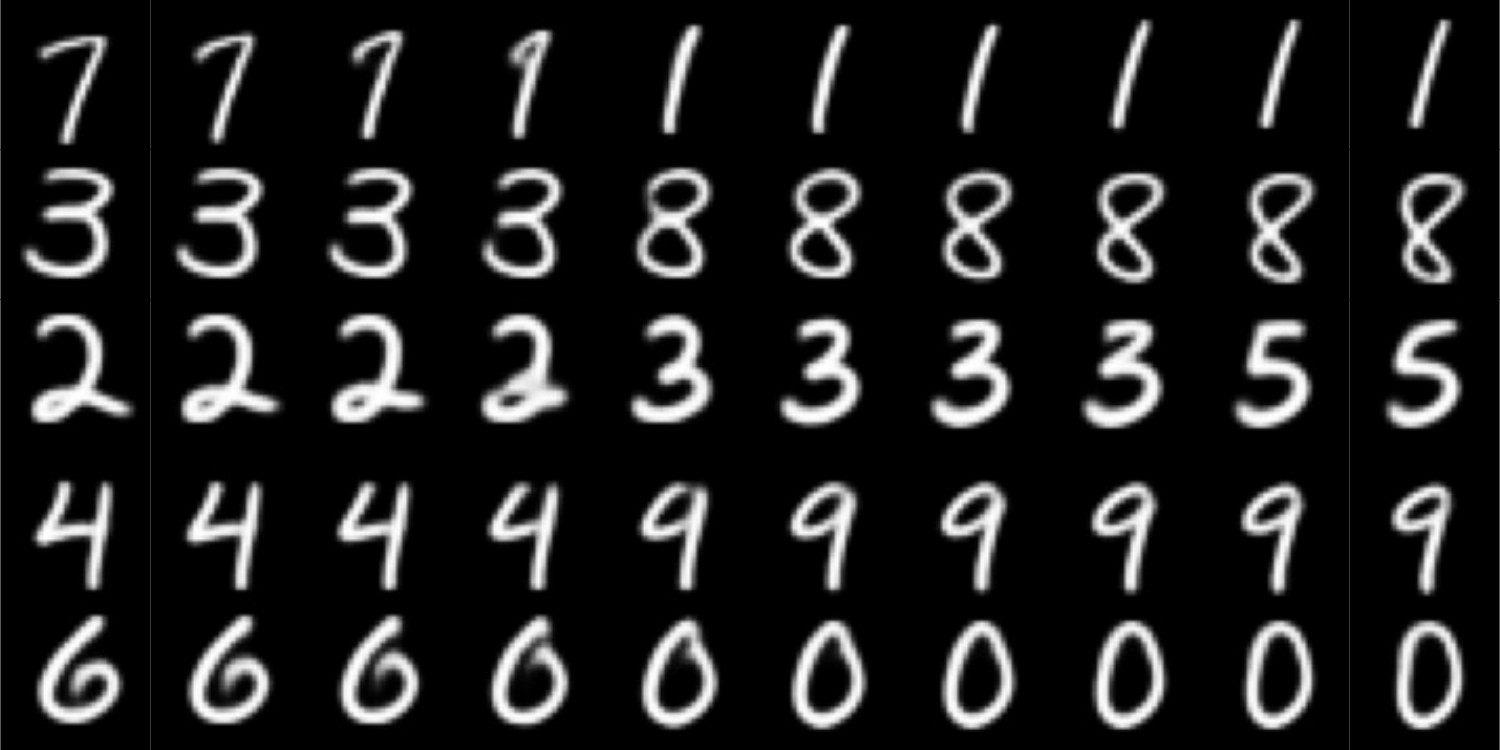}
         \caption{Unidirectional VAE}
         \label{fig:interp_mnist_vae}
   \end{subfigure}
   \vspace{0.1in}
      \hfill
   \begin{subfigure}[b]{0.24\linewidth}
         \centering
         \includegraphics[width=\textwidth]{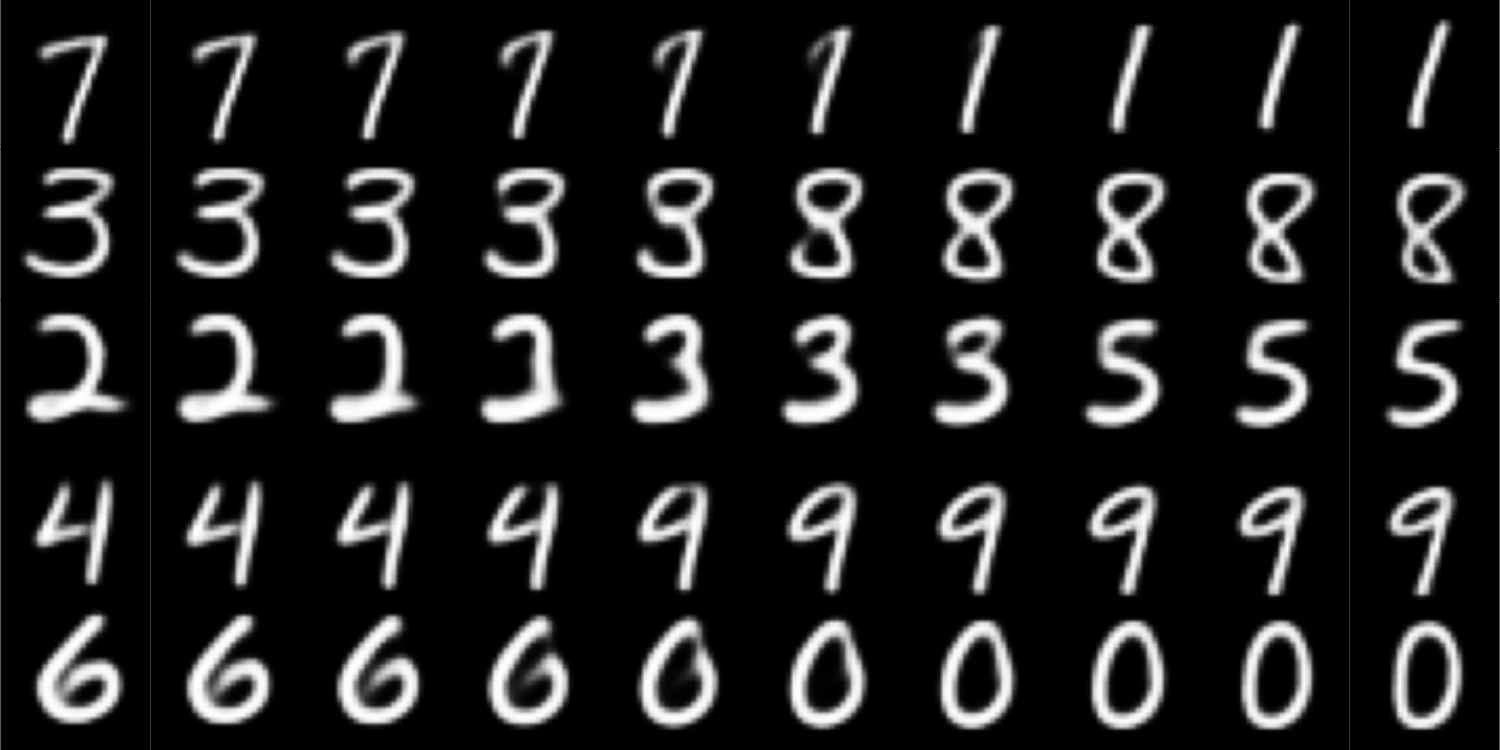}
         \caption{Bidirectional VAE (Ours)}
         \label{fig:interp_mnist_bvae}
   \end{subfigure}
         \hfill
\hfill
   \begin{subfigure}[b]{0.24\linewidth}
         \centering
         \includegraphics[width=\textwidth]{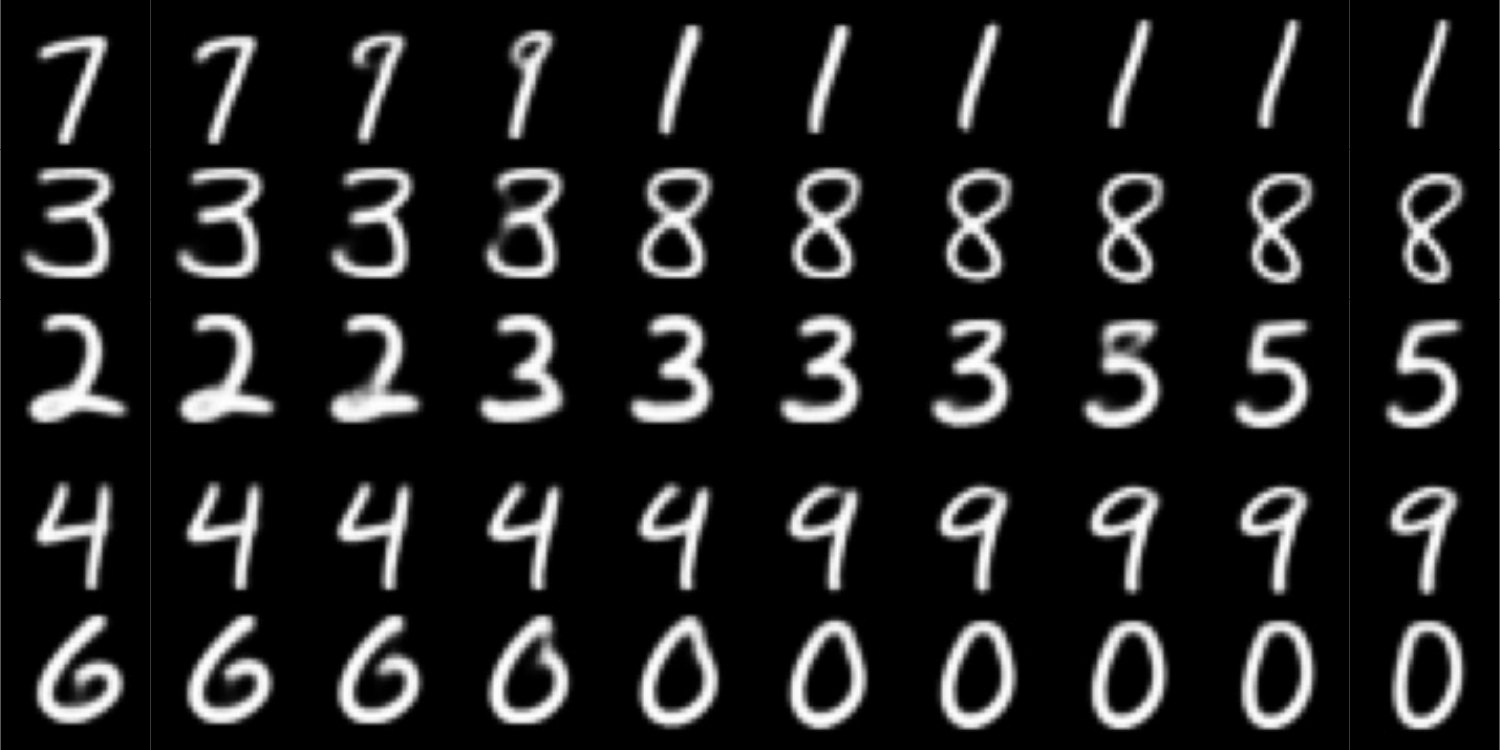}
         \caption{Unidirectional IWAE}
         \label{fig:interp_mnist_iwae}
   \end{subfigure}
\hfill
   \begin{subfigure}[b]{0.24\linewidth}
         \centering
         \includegraphics[width=\textwidth]{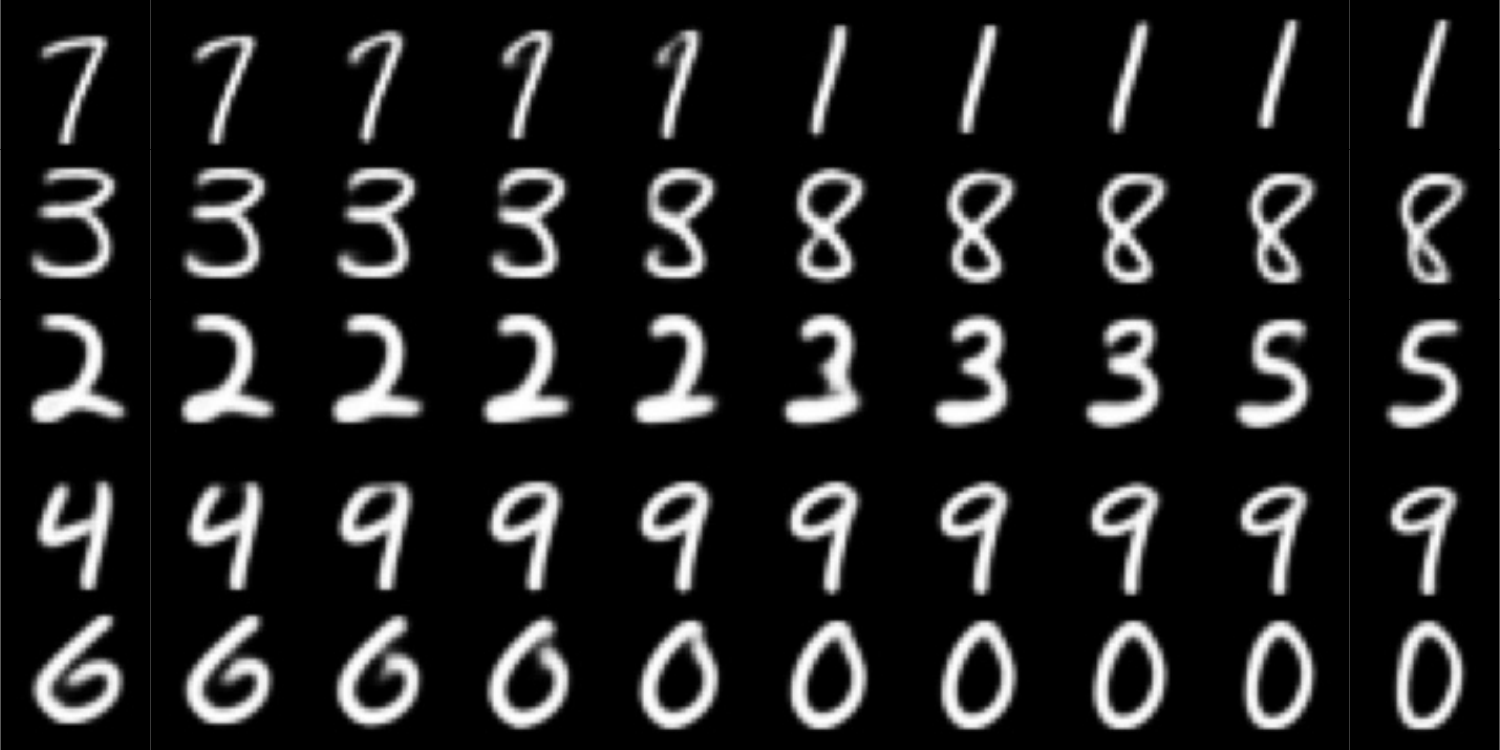}
         \caption{Bidirectional IWAE (Ours)}
         \label{fig:interp_mnist_biwae}
   \end{subfigure}
   \caption{\small MNIST handwritten image: Image interpolation with variational autoencoder networks.}
   \label{fig:mnist_interpolate}
\end{figure*}

\begin{table*}[!t]
  \centering
  \caption{\small Fashion-MNIST dataset with VAEs. We used the residual network architecture. The dimension of the latent variable is 64. 
  The BVAEs use $42.2$MB memory parameter and the unidirectional VAEs used $84$MB memory parameter.}
  \begin{tabular}{|l|c|ccc|cccc|}
    \hline
    \multicolumn{1}{|l|}{\multirow{3}{*}{\textbf{Model}}} & \multirow{3}{*}{\bf Parameters} & \multicolumn{3}{c|}{\multirow{2}{*}{\bf Generative Task}} & \multicolumn{4}{c|}{\multirow{2}{*}{ {\bf Reconstruction} and {\bf Classification}}} \\
    \multicolumn{1}{|l|}{\multirow{1}{*}}  &{} & \multicolumn{3}{|c|}{\multirow{2}{*}{}} & \multicolumn{4}{c|}{\multirow{2}{*}{}}\\
    {} & {}& {\bf NLL} $\downarrow$ & {\bf AU}$ \uparrow$ & {\bf FID}$\downarrow$  &  {\bf PSNR} $\uparrow$ & {\bf SSIM}$\uparrow$ &  {\bf rFID}$\downarrow$  & {\bf Accuracy} $\uparrow$ \\
    \hline
   VAE  & $22.1$M & ${231.3}$& {13} &{$3.082$}  & ${19.19}$ & ${ 0.6964}$& ${2.521}$ & {$87.17$\%} \\
    \textcolor[rgb]{0.,0.,0.9}{{Bidirectional} VAE} & \textcolor[rgb]{0,0,0.9}{{$11.1$M}}  & \textcolor[rgb]{0,0,0.9}{$231.1$}& \textcolor[rgb]{0,0,0.9}{\bf{$15$}}& \textcolor[rgb]{0,0,0.9}{$3.045$} &  \textcolor[rgb]{0,0,0.9}{$19.21$} & \textcolor[rgb]{0,0,0.9}{$0.6973$}& \textcolor[rgb]{0,0,0.9}{$2.478$} & \textcolor[rgb]{0,0,0.9}{\bf{$87.84\%$}} \\
        \hline
        \hline
   $\beta$-VAE ($\beta = 0.5$) & $22.1$M  & {$232.4$} & {$17$}& {$1.946$} &  {$20.01$} & {$0.7282$}& {$1.569$} & {$87.16\%$} \\
    \textcolor[rgb]{0.,0.,0.9}{{Bidirectional} $\beta$-VAE ($\beta = 0.5$) } & \textcolor[rgb]{0,0,0.9}{{$11.1$M}}  & \textcolor[rgb]{0,0,0.9}{{$232.4$}}& \textcolor[rgb]{0,0,0.9}{$20$} & \textcolor[rgb]{0,0,0.9}{$1.965$} &  \textcolor[rgb]{0,0,0.9}{$20.05$}& \textcolor[rgb]{0,0,0.9}{$0.7303$}& \textcolor[rgb]{0,0,0.9}{$1.658$} & \textcolor[rgb]{0,0,0.9}{${87.93\%}$}   \\
   \hline
   $\beta$-VAE ($\beta = 1.5$) & $22.1$M  & {$231.5$} & {$9$}& {$3.220$} &  {$18.59$} & {$0.6714$}& {$2.947$} & {$86.89\%$} \\
    \textcolor[rgb]{0.,0.,0.9}{{Bidirectional} $\beta$-VAE ($\beta = 1.5$) } & \textcolor[rgb]{0,0,0.9}{{$11.1$M}}  & \textcolor[rgb]{0,0,0.9}{{$231.5$}}& \textcolor[rgb]{0,0,0.9}{$9$} & \textcolor[rgb]{0,0,0.9}{$3.346$} &  \textcolor[rgb]{0,0,0.9}{$18.60$}& \textcolor[rgb]{0,0,0.9}{$0.6729$}& \textcolor[rgb]{0,0,0.9}{$2.893$} & \textcolor[rgb]{0,0,0.9}{${86.66\%}$}   \\
    \hline
    \hline
   $\beta$-TCVAE ($\beta = 0.5$) & $22.1$M  & {$232.3$} & {$24$}& {$2.420$} &  {$20.04$} & {$0.7238$}& {$1.996$} & {$87.71\%$} \\
    \textcolor[rgb]{0.,0.,0.9}{{Bidirectional} $\beta$-TCVAE ($\beta = 0.5$) } & \textcolor[rgb]{0,0,0.9}{{$11.1$M}}  & \textcolor[rgb]{0,0,0.9}{{$232.3$}}& \textcolor[rgb]{0,0,0.9}{$19$} & \textcolor[rgb]{0,0,0.9}{$2.026$} &  \textcolor[rgb]{0,0,0.9}{$20.09$}& \textcolor[rgb]{0,0,0.9}{$0.7288$}& \textcolor[rgb]{0,0,0.9}{$1.749$} & \textcolor[rgb]{0,0,0.9}{${88.06\%}$}   \\
    \hline
       $\beta$-TCVAE ($\beta = 1.0$) & $22.1$M  & {$231.1$} & {$12$}& {$2.833$} &  {$19.18$} & {$0.6941$}& {$2.427$} & {$87.21\%$} \\
    \textcolor[rgb]{0.,0.,0.9}{{Bidirectional} $\beta$-TCVAE ($\beta = 1.0$) } & \textcolor[rgb]{0,0,0.9}{{$11.1$M}}  & \textcolor[rgb]{0,0,0.9}{{$231.2$}}& \textcolor[rgb]{0,0,0.9}{$12$} & \textcolor[rgb]{0,0,0.9}{$2.891$} &  \textcolor[rgb]{0,0,0.9}{$19.21$}& \textcolor[rgb]{0,0,0.9}{$0.6954$}& \textcolor[rgb]{0,0,0.9}{$2.469$} & \textcolor[rgb]{0,0,0.9}{${87.06\%}$}   \\
     \hline
       $\beta$-TCVAE ($\beta = 1.5$) & $22.1$M  & {$231.8$} & {$9$}& {$3.579$} &  {$18.54$} & {$0.6678$}& {$3.234$} & {$86.62\%$} \\
    \textcolor[rgb]{0.,0.,0.9}{{Bidirectional} $\beta$-TCVAE ($\beta = 1.5$) } & \textcolor[rgb]{0,0,0.9}{{$11.1$M}}  & \textcolor[rgb]{0,0,0.9}{{$231.4$}}& \textcolor[rgb]{0,0,0.9}{$10$} & \textcolor[rgb]{0,0,0.9}{$3.273$} &  \textcolor[rgb]{0,0,0.9}{$18.58$}& \textcolor[rgb]{0,0,0.9}{$0.6702$}& \textcolor[rgb]{0,0,0.9}{$2.969$} & \textcolor[rgb]{0,0,0.9}{${87.00\%}$}   \\
    \hline
    \hline
   IWAE  & $22.1$M & ${230.3}$& {$17$} &{$2.489$}  & ${17.78}$ & ${0.6529}$& ${2.070}$ & {$88.14\%$} \\
    \textcolor[rgb]{0.,0.,0.9}{{Bidirectional} IWAE}  & \textcolor[rgb]{0,0,0.9}{{$11.1$M}}  & \textcolor[rgb]{0,0,0.9}{$230.4$}& \textcolor[rgb]{0,0,0.9}{\bf{$14$}}& \textcolor[rgb]{0,0,0.9}{$2.881$} &  \textcolor[rgb]{0,0,0.9}{$18.10$} & \textcolor[rgb]{0,0,0.9}{$0.6559$}& \textcolor[rgb]{0,0,0.9}{$2.432$} & \textcolor[rgb]{0,0,0.9}{\bf{$88.05\%$}} \\
    \hline
  \end{tabular}
  \label{tab:fmnist_latent_dim_64_residual_network}
\end{table*}

\begin{figure*}[t]
   \begin{subfigure}[b]{0.24\linewidth}
         \centering
         \includegraphics[width=\textwidth]{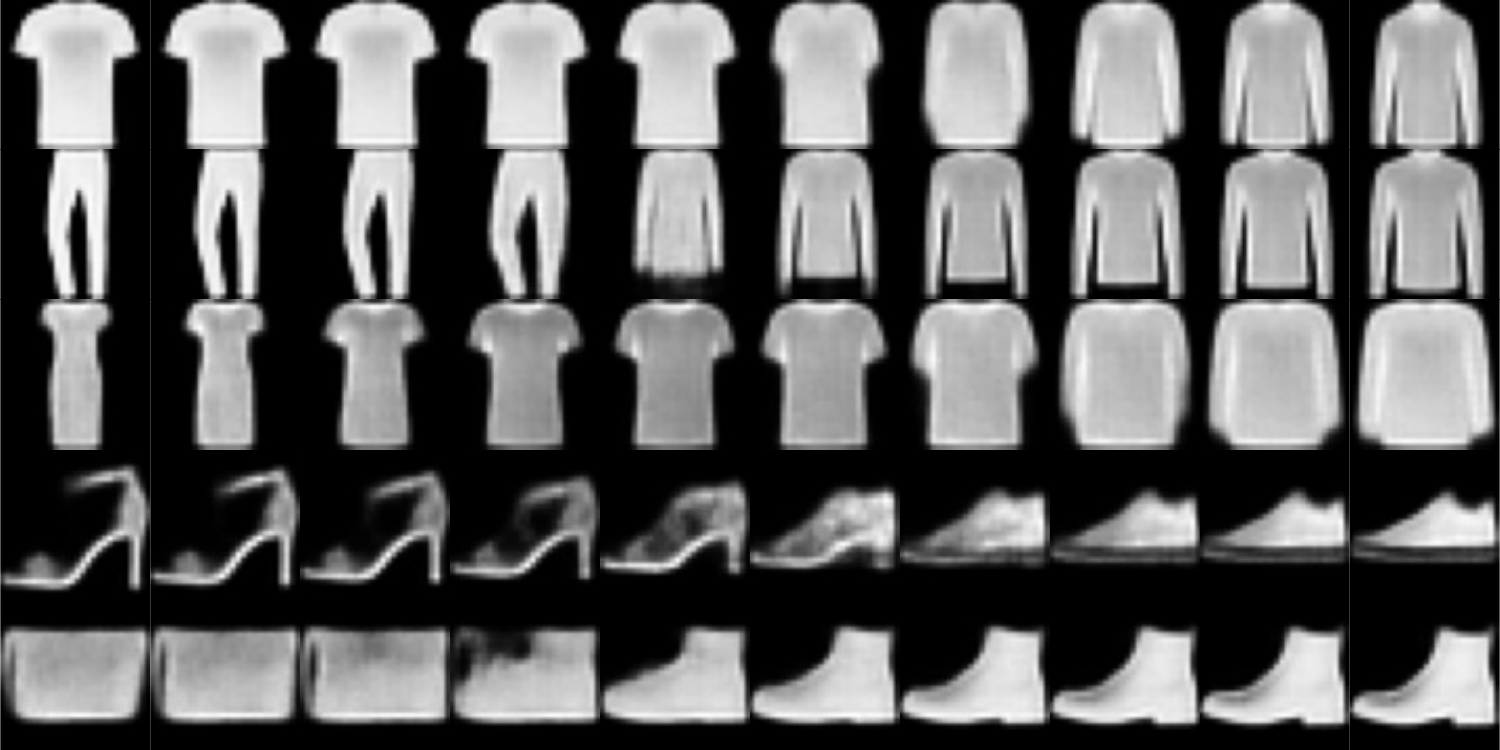}
         \caption{Unidirectional VAE}
         \label{fig:interp_fmnist_vae}
   \end{subfigure}
   \vspace{0.1in}
      \hfill
   \begin{subfigure}[b]{0.24\linewidth}
         \centering
         \includegraphics[width=\textwidth]{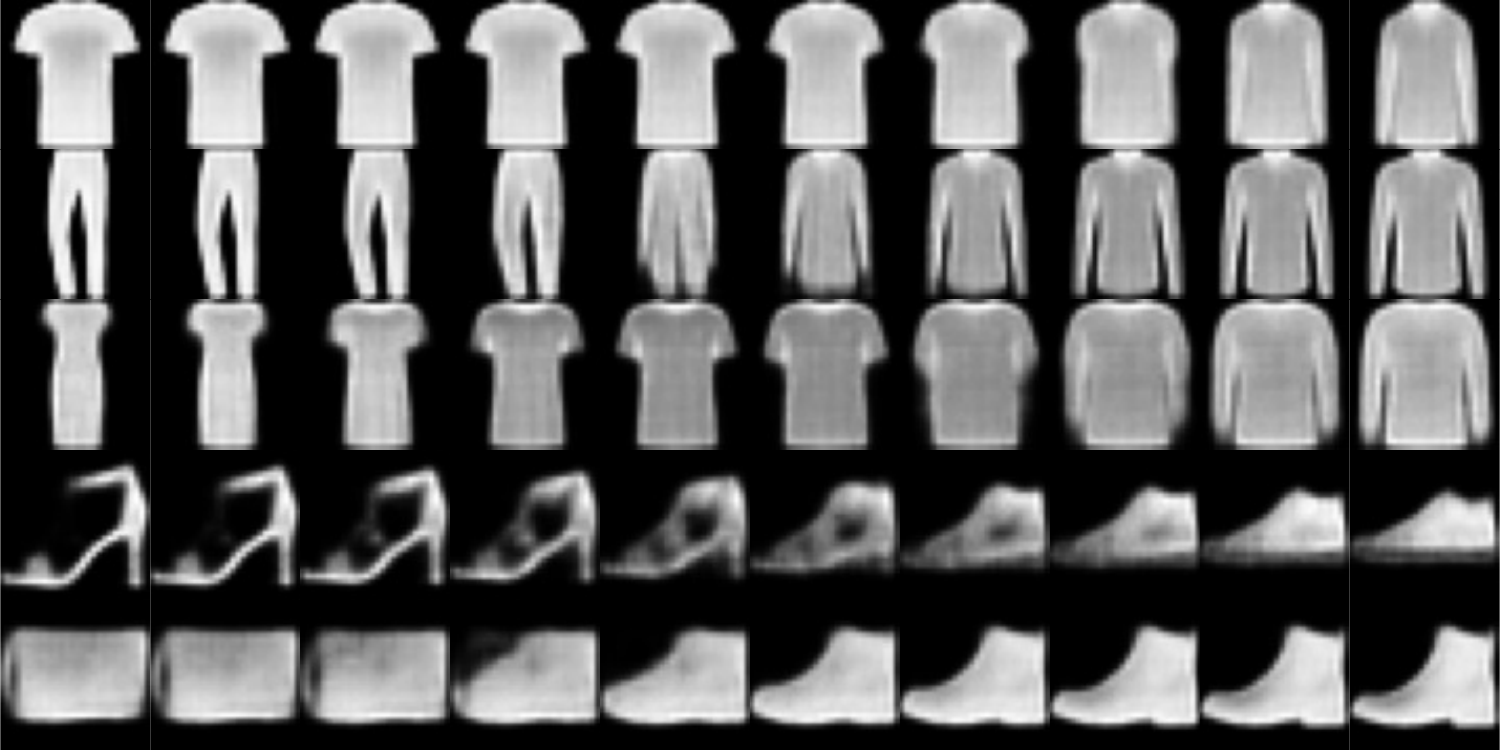}
         \caption{Bidirectional VAE (Ours)}
         \label{fig:interp_fmnist_bvae}
   \end{subfigure}
         \hfill
   \begin{subfigure}[b]{0.24\linewidth}
         \centering
         \includegraphics[width=\textwidth]{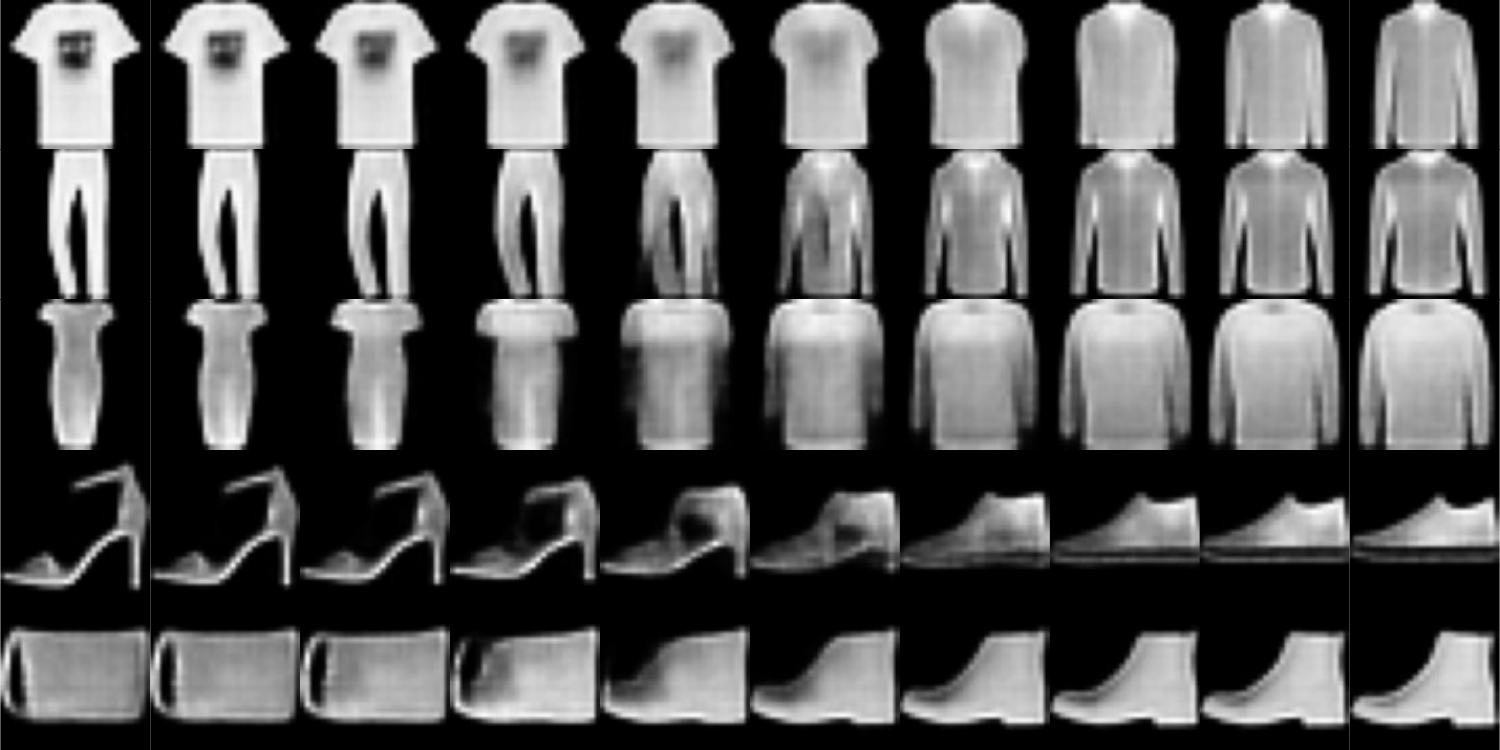}
         \caption{Unidirectional $\beta$-VAE}
         \label{fig:interp_fmnist_beta_vae}
   \end{subfigure}
            \hfill
   \begin{subfigure}[b]{0.24\linewidth}
         \centering
         \includegraphics[width=\textwidth]{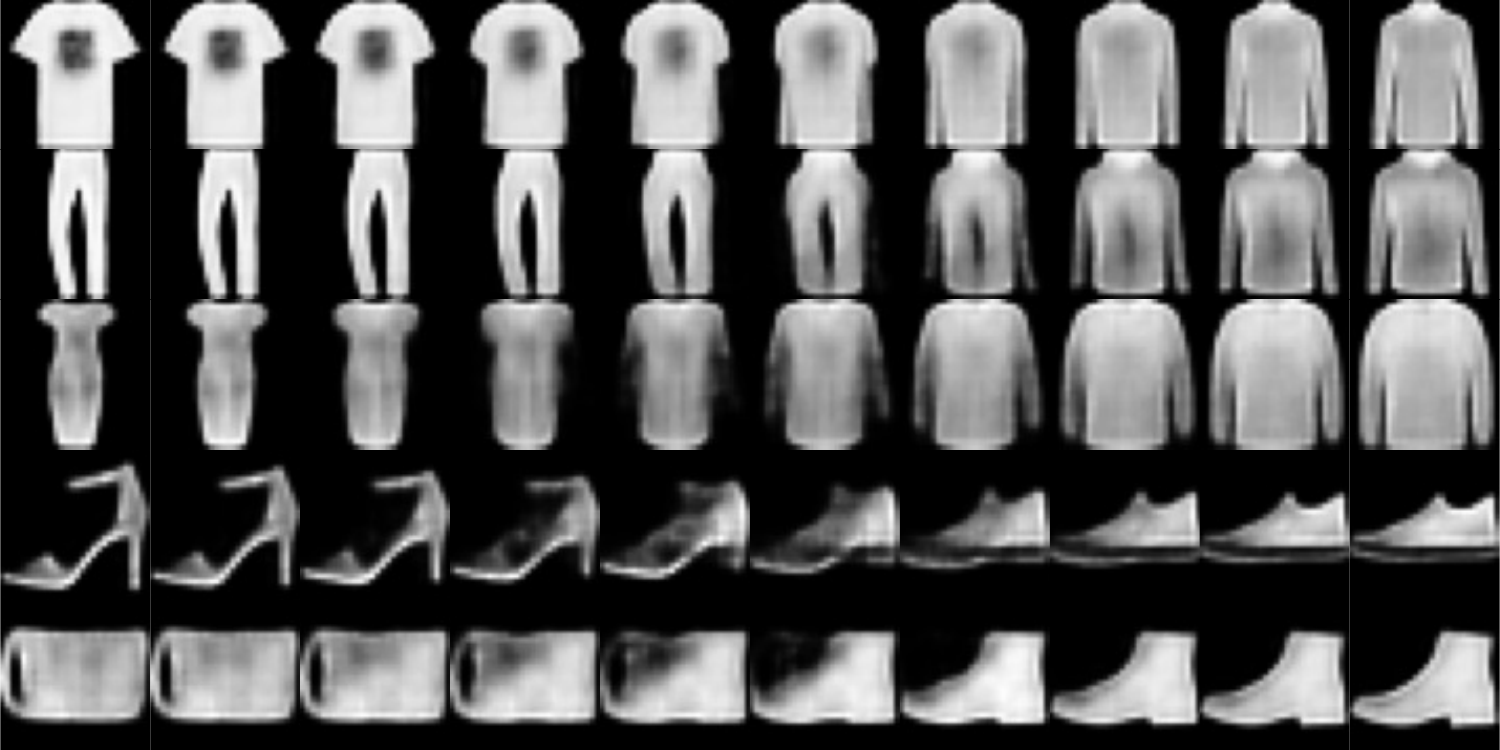}
         \caption{Bidirectional $\beta$-VAE (Ours)}
         \label{fig:interp_fmnist_beta_bvae}
   \end{subfigure}
   \caption{\small{Image interpolation with VAEs on the Fashion-MNIST dataset.}}
   \label{fig:mnist_interpolate}
\end{figure*}

\textit {Downstream Classification}: We trained simple linear classifiers on VAE-extracted features from the MNIST digit dataset. 
 We trained shallow neural classifiers with one hidden layer and 100 hidden neurons each on the extracted features from the Fashion-MNIST images.
Similar neural classifiers with one hidden layer and 256 hidden neurons each trained on VAE-extracted features from the CIFAR-10 dataset.

\subsection{Training}
We considered four implementations of VAEs and compare them with their respective bidirectional versions.
The four VAEs are vanilla VAE {\cite{KingmaW13}, $\beta$-VAE \cite{higgins2016beta}, $\beta$-TCVAE  \cite{chen2018isolating}, and IWAE \cite{burda2015importance}.
We trained these over the four datasets across the four tasks.
We used the AdamW optimizer \cite{loshchilov2017decoupled} with the OneCycleLR \cite{smith2019super} learning rate scheduler.
The optimizer trained on their respective ELBO estimates. 

We designed a new framework for bidirectional VAEs and implemented unidirectional VAEs with the Pythae framework \cite{ChadebecVA22}.
All the models trained on a single A100 GPU.
Tables \ref{tab:mnist_latent_dim_64} - \ref{tab:celeb_a} and Figures \ref{fig:tsne} - \ref{fig:mnist_interpolate} present the results.

\begin{table*}[!t]
  \centering
  \caption{\small CIFAR-10 dataset with VAEs. The dimension of the latent space is $256$.
  The BVAEs each used $107$MB memory parameter and the unidirectional VAEs each used $214$MB memory parameter.}
  \begin{tabular}{|l|c|ccc|cccc|}
    \hline
    \multicolumn{1}{|l|}{\multirow{3}{*}{\textbf{Model}}} & \multirow{3}{*}{\bf Parameters} & \multicolumn{3}{c|}{\multirow{2}{*}{\bf Generative Task}} & \multicolumn{4}{c|}{\multirow{2}{*}{ {\bf Reconstruction} and {\bf Classification}}} \\
    \multicolumn{1}{|l|}{\multirow{1}{*}}  &{} & \multicolumn{3}{c|}{\multirow{2}{*}{}} & \multicolumn{4}{c|}{\multirow{2}{*}{}}\\
    {} & {}& {\bf NLL} $\downarrow$ & {\bf AU}$\uparrow$ & {\bf FID}$\downarrow$  &  {\bf PSNR} $\uparrow$ & {\bf SSIM}$\uparrow$ &  {\bf rFID}$\downarrow$  & {\bf Accuracy} $\uparrow$ \\
   \hline
    VAE  & $56.6$M & ${1817.0}$ & ${51}$ & $2.483$ & ${18.67}$ & ${0.4532}$ &  ${2.305}$ & {$51.73\%$} \\
    \textcolor[rgb]{0,0,0.9}{{Bidirectional} VAE} & \textcolor[rgb]{0,0,0.9}{{$28.3$M}} & \textcolor[rgb]{0,0,0.9}{{$1814.1$}}  & \textcolor[rgb]{0,0,0.9}{{$37$}} &   \textcolor[rgb]{0,0,0.9}{{$2.442$}} & \textcolor[rgb]{0,0,0.9}{{$18.93$}}& \textcolor[rgb]{0,0,0.9}{{$0.4742$}} & \textcolor[rgb]{0,0,0.9}{{$2.255$}} & \textcolor[rgb]{0,0,0.9}{\bf{$50.55\%$}}  \\
        \hline
        \hline
    $\beta$-VAE ($\beta = 0.5$) & $56.6$M & ${1816.7}$ & ${85}$ & ${2.225} $ & ${19.48}$ & ${0.5191}$ &  ${1.955}$  & {$51.78\%$} \\
    \textcolor[rgb]{0,0,0.9}{{Bidirectional} $\beta$-VAE ($\beta = 0.5$) } & \textcolor[rgb]{0,0,0.9}{{$28.3$M}} & \textcolor[rgb]{0,0,0.9}{{$1815.1$}}  & \textcolor[rgb]{0,0,0.9}{$65$} & \textcolor[rgb]{0,0,0.9}{{$2.122$}} & \textcolor[rgb]{0,0,0.9}{${19.83}$}& \textcolor[rgb]{0,0,0.9}{{$0.5466$}} & \textcolor[rgb]{0,0,0.9}{{$1.827$}} & \textcolor[rgb]{0,0,0.9}{{$51.74\%$}}  \\
            \hline
    $\beta$-VAE ($\beta = 1.5$) & $56.6$M & ${1820.9}$  & $32$ & ${2.625} $ & $18.12$ & $0.4113$ & $2.483$ & $50.44\%$\\
    \textcolor[rgb]{0.0,0,0.9}{{Bidirectional} $\beta$-VAE ($\beta = 1.5$)} & \textcolor[rgb]{0.0,0,0.9}{{$28.3$M}}& \textcolor[rgb]{0.0,0,0.9}{{$1816.1$}}& \textcolor[rgb]{0.0,0,0.9}{${40}$} & \textcolor[rgb]{0,0,0.9}{{$2.535$}} & \textcolor[rgb]{0.0,0,0.9}{${18.39}$} & \textcolor[rgb]{0.0,0,0.9}{{$0.4337$}} &  \textcolor[rgb]{0.0,0,0.9}{{$2.342$}} & \textcolor[rgb]{0.0,0,0.9}{\bf{$51.91\%$}} \\
    \hline
   \hline
    $\beta$-TCVAE ($\beta = 0.5$) & $56.6$M & ${1816.8}$  & ${50}$ & ${2.225}$ & $19.26$ & $0.5035$ & $2.035$ & $50.96$\%\\
    \textcolor[rgb]{0.0,0,0.9}{{Bidirectional} $\beta$-TCVAE ($\beta = 0.5$)} & \textcolor[rgb]{0.0,0,0.9}{${28.8}$M}& \textcolor[rgb]{0.0,0,0.9}{${1814.0}$}& \textcolor[rgb]{0.0,0,0.9}{$59$} & \textcolor[rgb]{0,0,0.9}{{$2.101$}} & \textcolor[rgb]{0.0,0,0.9}{${19.63}$} & \textcolor[rgb]{0.0,0,0.9}{${0.5321}$} &  \textcolor[rgb]{0.0,0,0.9}{${1.942}$} & \textcolor[rgb]{0.0,0,0.9}{${50.51\%}$} \\
        \hline
    $\beta$-TCVAE ($\beta = 1.0$) & $56.6$M & ${1819.6}$  & $27$ & $ 2.485$ & $18.47$ & $0.4362$ & $2.348$ & $49.57\%$\\
    \textcolor[rgb]{0.0,0,0.9}{{Bidirectional} $\beta$-TCVAE ($\beta = 1.0$)} & \textcolor[rgb]{0.0,0,0.9}{${28.3}$M}& \textcolor[rgb]{0.0,0,0.9}{${1814.0}$}& \textcolor[rgb]{0.0,0,0.9}{${42}$} & \textcolor[rgb]{0,0,0.9}{{$2.380$}} & \textcolor[rgb]{0.0,0,0.9}{${18.91}$} & \textcolor[rgb]{0.0,0,0.9}{${0.4752}$} &  \textcolor[rgb]{0.0,0,0.9}{${2.201}$} & \textcolor[rgb]{0.0,0,0.9}{${50.93}\%$} \\
  \hline
    $\beta$-TCVAE ($\beta = 1.5$) & $56.6$M & ${1824.1}$  & $33$ & $2.673 $ & $17.93$ & $0.3940$ &  $2.604$ & $50.52\%$\\
    \textcolor[rgb]{0.0,0,0.9}{{Bidirectional} $\beta$-TCVAE ($\beta = 1.5$) } & \textcolor[rgb]{0.0,0,0.9}{${28.3}$M} & \textcolor[rgb]{0.0,0,0.9}{${1816.8}$}& \textcolor[rgb]{0.0,0,0.9}{${35}$} & \textcolor[rgb]{0,0,0.9}{{$2.576$}} & \textcolor[rgb]{0.0,0,0.9}{${18.36}$} & \textcolor[rgb]{0.0,0,0.9}{${0.4315}$} &  \textcolor[rgb]{0.0,0,0.9}{${2.437}$} & \textcolor[rgb]{0.0,0,0.9}{${51.04\%}$} \\
   \hline
    \hline
    IWAE   & $56.6$M & ${1814.6}$  &  $122$  & $2.415$ & $18.55$ & $0.4540$ & $2.173$  & $51.94\%$ \\
    \textcolor[rgb]{0.0,0,0.9}{{Bidirectional} IWAE}& \textcolor[rgb]{0.0,0,0.9}{${28.3}$M}& \textcolor[rgb]{0.0,0,0.9}{${1811.5}$}  & \textcolor[rgb]{0.0,0,0.9}{${128}$} & \textcolor[rgb]{0,0,0.9}{{$2.309$}}& \textcolor[rgb]{0.0,0,0.9}{${18.84}$} & \textcolor[rgb]{0.0,0,0.9}{${0.4838}$} & \textcolor[rgb]{0.0,0,0.9}{${2.015}$} & \textcolor[rgb]{0.0,0,0.9}{${51.65\%}$}  \\
 \hline
  \end{tabular}
  \label{tab:cifar_10_bvaes}
\end{table*}

\begin{table*}
\centering
  \caption{CelebA-64  dataset with VAEs. The dimension of the latent variable is $256$. The BVAEs each used $133.8$MB memory parameter and the unidirectional VAEs each used $267.6$MB memory parameter.}
  \begin{tabular}{|l|c|cc|cc|}
   \hline
    \multicolumn{1}{|l|}{\multirow{3}{*}{\textbf{Model}}} & \multirow{3}{*}{\bf Parameters} & \multicolumn{2}{|c|}{\multirow{2}{*}{\bf Generative Task}} & \multicolumn{2}{c|}{\multirow{2}{*}{ {\bf Reconstruction} }} \\
    \multicolumn{1}{|l|}{\multirow{1}{*}}  &{} & \multicolumn{2}{|c|}{\multirow{2}{*}{}} & \multicolumn{2}{c|}{\multirow{2}{*}{}}\\
    {} & {}& {\bf NLL} $\downarrow$ & {\bf AU}$\uparrow$ &  {\bf PSNR} $\uparrow$ & {\bf SSIM}$\uparrow$ \\
    \hline
   VAE & {$69.1$M} & {$6221.9$}& {$80$} & {$20.54$}& {$0.6528$}  \\
    \textcolor[rgb]{0.0,0,0.9}{{Bidirectional} VAE} & \textcolor[rgb]{0.0,0,0.9}{{$34.6$M}}  & \textcolor[rgb]{0.0,0,0.9}{{$6217.7$}} & \textcolor[rgb]{0.0,0,0.9}{$52$} & \textcolor[rgb]{0.0,0,0.9}{{$20.66$}}& \textcolor[rgb]{0.0,0,0.9}{{$0.6532$}}  \\
        \hline
   $\beta$-VAE ($\beta=0.1$)  & $69.1$M & {$6243.0$}& {$240$} & {$22.35$}& {$0.7225$}  \\
    \textcolor[rgb]{0.0,0,0.9}{{Bidirectional} $\beta$-VAE} & \textcolor[rgb]{0.0,0,0.9}{{$34.6$M}}  & \textcolor[rgb]{0.0,0,0.9}{{$6261.9$}}&\textcolor[rgb]{0.0,0,0.9}{$90$} & \textcolor[rgb]{0.0,0,0.9}{{$22.60$}} & \textcolor[rgb]{0.0,0,0.9}{{$0.7243$}}    \\
        \hline
   $\beta$-TCVAE ($\beta = 0.1$)  & $69.1$M & ${6277.2}$& {$93$} & ${22.15}$ & ${0.7189}$  \\
    \textcolor[rgb]{0.0,0,0.9}{{Bidirectional} $\beta$-TCVAE} & \textcolor[rgb]{0.0,0,0.9}{{$34.6$M}}  & \textcolor[rgb]{0.0,0,0.9}{{$6275.2$}}& \textcolor[rgb]{0.0,0,0.9}{{$145$}} & \textcolor[rgb]{0.0,0,0.9}{{$23.55$}} & \textcolor[rgb]{0.0,0,0.9}{{$0.7681$}}    \\
   \hline
   $\beta$-IWAE  & $69.1$M & {$6221.1$}& {$77$} & {$20.44$}& {$0.6521$}  \\
    \textcolor[rgb]{0.0,0,0.9}{{Bidirectional} $\beta$-IWAE} & \textcolor[rgb]{0.0,0,0.9}{{$34.6$M}}  & \textcolor[rgb]{0.0,0,0.9}{{$6217.5$}}& \textcolor[rgb]{0.0,0,0.9}{{$204$}} & \textcolor[rgb]{0.0,0,0.9}{{$20.81$}} & \textcolor[rgb]{0.0,0,0.9}{{$0.6551$}}    \\
    \hline
  \end{tabular}
  \label{tab:celeb_a}
\end{table*}

\subsection{Evaluation Metrics}

We measured how the VAE models performed on generative and compression tasks.
We used these 6 quantitative metrics:

\begin{itemize}
  \item \textit{Negative Log-Likelihood (NLL)}: NLL estimates the negative of $\ln p(x|\theta)$.
  This is computationally intractable and so we used Monte Carlo approximation.
  A lower value means that the model generalizes well on unseen data.
  \item \textit{Number of Active Latent Units (AU)} \cite{BurdaGS15}:  This metric reflects the number of latent variables $\bf z$ that have a variance above a given threshold $\epsilon$:
  \begin{align}
  AU &= \sum_{d=1}^{D} \mathbb{I}\Big{[} \mathit{Cov}_{\bf x}\big{[} \mathbb{E}_{{\bf z}|{\bf x},\phi}[{\bf z}_d] \big{]} \geq \epsilon \Big{]}
  \end{align}  
  where $\mathbb{I}$ is an indicator function, ${\bf z}_d$ represents the $d^{th}$ component of the latent variable $\bf z$, and $\epsilon = 0.01$.
 A higher AU means that the model uses more features to represent the latent space.
Too many active units can lead to overfitting.

 \item \textit{Peak Signal-to-Noise Ratio (PSNR)}: This common metric compares the reconstructed images with their  target images.
 A higher value implies a better reconstruction from data compression.
 \item \textit{Structural Similarity Index (SSIM)}: SSIM is a perceptual metric.
 It quantifies how the data compression degrades the image.
 A higher SSIM value implies a better reconstruction from image compression.
 \item \textit{Downstream Classification Accuracy}: This metric is just the classification accuracy of simple classifiers that trained on the latent or VAE-extracted features.
 A higher accuracy means that the compression extracts easy-to-classify features. 
 \item \textit{Fr\'echet Inception Distance (FID)}\cite{HeuselRUNH17}: This metric evaluates the quality of generated images.
It measures the similarity between the distribution of the real images and the distribution of the generated images. 
 A lower value implies that the generated images are closer to the real images. 
\end{itemize}

\section{Conclusion}

Bidirectional VAEs encode and decode through the same synaptic web of a deep neural network.
This bidirectional flow captures the joint probabilistic structure of both directions during learning and recall.
They cut the synaptic parameter count roughly in half the parameter compared with unidirectional VAEs.
The simulations on the four image test sets showed that the bidirectional VAEs still performed slightly better than the unidirectional VAEs.

\bibliographystyle{IEEEtrans}
\bibliography{vae}


\end{document}